\documentclass[lettersize,journal]{IEEEtran}
\usepackage{amsmath,amsfonts}
\usepackage{array}
\usepackage[caption=false,font=normalsize,labelfont=sf,textfont=sf]{subfig}
\usepackage{textcomp}
\usepackage{stfloats}
\usepackage{url}
\usepackage{verbatim}
\usepackage{graphicx}

\usepackage[table,xcdraw]{xcolor}
\usepackage{tabularx}
\usepackage{makecell}
\usepackage{multirow}
\usepackage{booktabs}
\usepackage{bm}
\usepackage{color}
\usepackage{balance}
\usepackage{xcolor}
\usepackage{algorithm}
\usepackage{amssymb}

\usepackage{algpseudocode} 

\newcommand{\newtext}[1]{{\leavevmode\color{black}#1}}

\newlength{\bioPhotoWidth}   
\newlength{\bioPhotoHeight}  
\usepackage{relsize}   

\def\BibTeX{{\rm B\kern-.05em{\sc i\kern-.025em b}\kern-.08em
    T\kern-.1667em\lower.7ex\hbox{E}\kern-.125emX}}
    
\begin{document}
\title{Hierarchical Anti-Aesthetics: Protecting Facial Privacy against Customized Diffusion Models}

\author{Songping Wang, Yueming Lyu$^\dagger$, Shiqi Liu, Chen Zhao, Ziyuan Chen, Ning Li, Jing Dong, Senior Member, IEEE, Caifeng Shan$^\dagger$, Senior Member, IEEE
\thanks{Songping Wang, Yueming Lyu, Shiqi Liu, Chen Zhao, Ziyuan Chen, Caifeng Shan are with the School of Intelligence Science and Technology, Nanjing University, Suzhou 215163, China (email: \texttt{theone@buaa.edu.cn}).}
\thanks{Ning Li is with the China Mobile Information Technology Co., Ltd., Beijing 100037, China.}
\thanks{Jing Dong is with the State Key Laboratory of Multimodal Artificial Intelligence Systems (MAIS), Center for Research on Intelligent Perception and Computing (CRIPAC), Institute of Automation Chinese Academy of Sciences (CASIA), Beijing 100190, China.}
\thanks{$^\dagger$Corresponding author}
}

\markboth{MANUSCRIPT FOR IEEE TRANSACTIONS ON INFORMATION FORENSICS AND SECURITY}%
{How to Use the IEEEtran \LaTeX \ Templates}

\maketitle

\begin{abstract}

The rise of customized diffusion models has fueled a boom in personalized visual content creation, but it also introduces serious risks of malicious misuse, thereby posing threats to personal privacy. Image aesthetics are strongly correlated with human perception of image quality. Motivated by this observation, we address facial privacy protection from a novel aesthetic perspective by degrading the generation quality of maliciously customized models, thus reducing facial identity leakage. Specifically, we propose a Hierarchical Anti-Aesthetics (HAA) framework that exploits aesthetic cues at multiple perceptual levels. HAA consists of two key branches: (1) Global Anti-Aesthetics, which degrades overall aesthetics and generation quality by constructing a global anti-aesthetic reward mechanism and a corresponding loss; and (2) Local Anti-Aesthetics, which disrupts facial identity by using a local anti-aesthetic reward mechanism and loss to guide adversarial perturbations toward facial regions. By integrating both branches, HAA achieves anti-aesthetic degradation from a global to a local level during customized generation. Extensive experiments show that HAA outperforms existing methods in identity removal, providing an effective tool for protecting facial privacy.

\end{abstract}


\section{Introduction}
Deep learning, as a core technology in the field of artificial intelligence, has developed rapidly in recent years and provided important support for a wide range of vision tasks~\cite{wang2024effective,meng2023coarse,chen2026neurerase,jin2025frequency,zhou2025multimodal,wang2025fast,wang2026exposing,wang2024public}. Diffusion models (DMs), as an important class of deep generative models, have achieved significant breakthroughs in text-to-image (T2I) generation~\cite{gu2022vector,rombach2022high,ramesh2022hierarchical,ho2020denoising,song2020denoising,bar2022text2live,kim2022diffusionclip}. These models diversify and enhance visual effects in image generation while ensuring a high degree of consistency between the generated images and textual descriptions. To meet personalized needs and improve fine-tuning efficiency, researchers have developed various DM fine-tuning methods, such as Textual Inversion~\cite{gal2022image}, DreamBooth~\cite{ruiz2023DreamBooth}, Custom Diffusion~\cite{kumari2023multi}, and SVDiff~\cite{han2023svdiff}. These fine-tuning methods offer stronger customization capabilities, allowing users to generate high-quality images of specific themes with few reference images.

\begin{figure}[t]
  \centering
  \includegraphics[width=\linewidth]{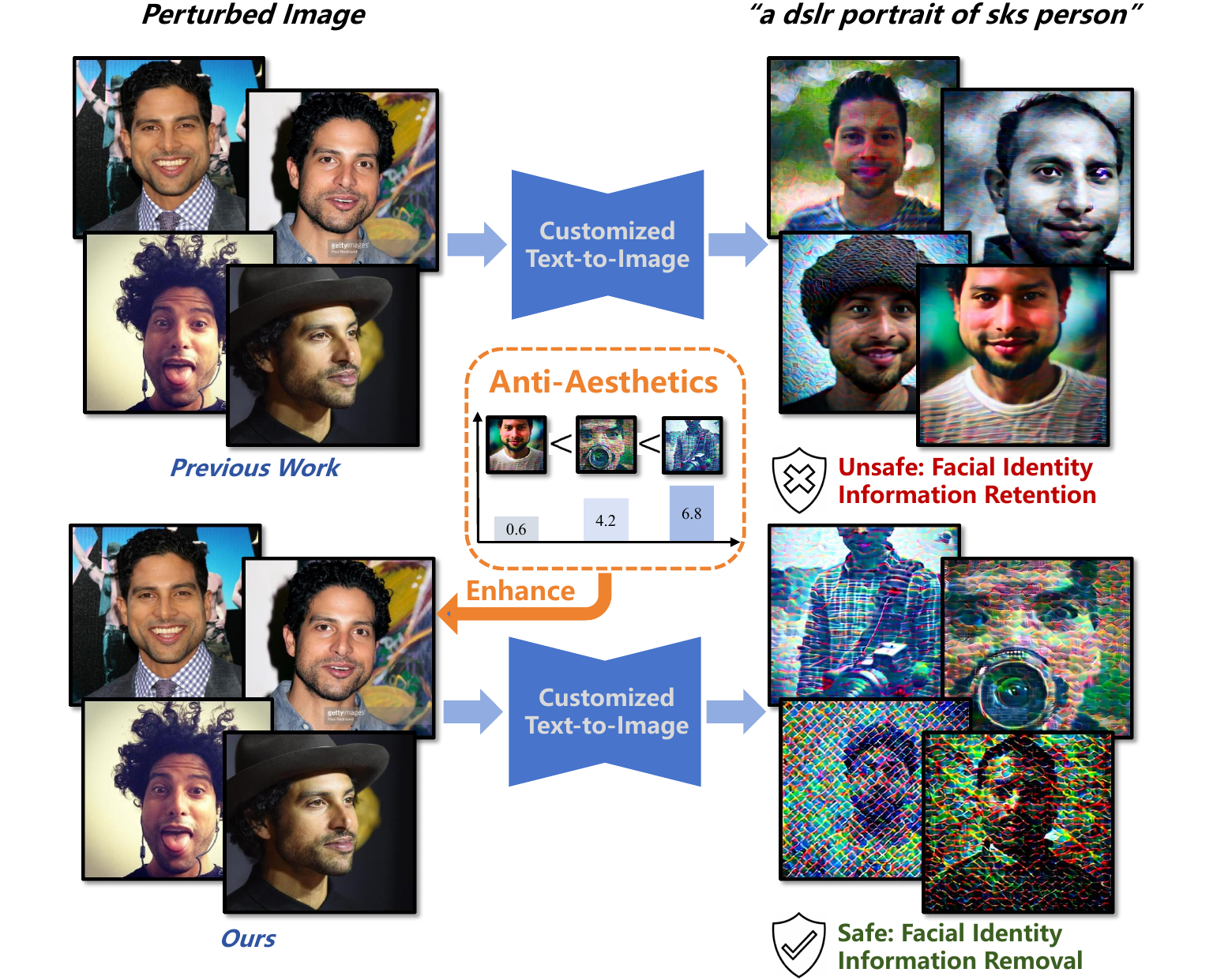}
  \caption{\newtext{Previous image-protection methods overlook aesthetic cues, which limits their ability to remove identity and inevitably leads to privacy leakage.} In contrast, our method effectively enhances the ability to eliminate facial identity, guided by the proposed anti-aesthetic mechanisms. The prompt is ``a dslr portrait of sks person''.}
  \label{fig:motivation}
  \vspace{-10pt}
\end{figure}

\begin{figure*}[t]
  \centering
  \includegraphics[width=\textwidth]{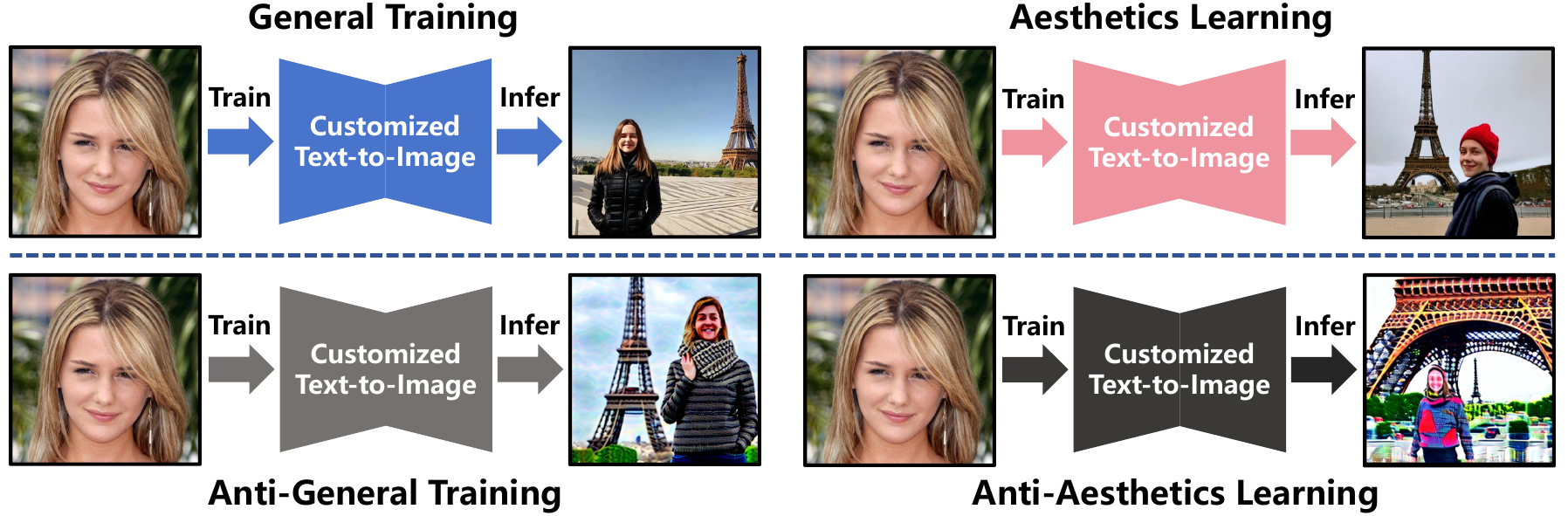}
  \caption{Compared with general training, aesthetic learning enhances the quality and details of generated images by aligning with human aesthetic preferences. Similarly, compared with anti-general training (previous protection methods), anti-aesthetic learning further reduces the quality of generated images and destroys facial details by misaligning with human aesthetics, thereby effectively protecting facial privacy. ``a photo of a sks person in front of the Eiffel Tower'' is used as the prompt for inference.}
  \label{fig:obs}
  \vspace{-10pt}
\end{figure*}

Despite the immense potential of these technologies, they also harbor significant safety risks that cannot be ignored~\cite{wang2026runawayevil,wang2025exploring,wang2025anti,wei2023efficient,li2026agentcanary}. Malicious users may exploit these models to generate forged images or deepfake content, infringing on personal privacy and intellectual property rights, and even creating fake news to mislead the public~\cite{vice2024bagm,huang2024personalization,frank2020leveraging,seow2022comprehensive}.

To address these threats, anti-customization methods are primarily based on adversarial attacks, such as Mist~\cite{zheng2023understanding}, ASPL~\cite{van2023anti}, and CAAT~\cite{xu2024perturbing}. These methods introduce adversarial noise to interfere with the customized fine-tuning process, preventing the malicious misuse of user images by customized diffusion models. However, as illustrated in Fig.~\ref{fig:motivation}, existing methods exhibit limitations in facial privacy protection and copyright preservation. These approaches primarily rely on a straightforward end-to-end paradigm that degrades overall image quality by maximizing the original training loss, yet they critically overlook the design of identity elimination for local facial regions. This oversight substantially undermines their effectiveness in removing recognizable facial features, leading to privacy leakage and the misuse of user portraits. \newtext{Recent methods refine this pipeline by attacking cross-attention, timestep/frequency features, or global-local feature/attribute constraints~\cite{xu2024perturbing,wang2024simac,xu2025harnessing}. In contrast, HAA introduces a preference-level view of protection: malicious customization is weakened by suppressing the human-preferred realism and facial details required for usable identity reconstruction. Under the same constrained perturbation interface, frozen reward models impose explicit anti-aesthetic targets at whole-image and face-local levels, providing a direct signal beyond model-internal feature disruption.}

To delve into this issue, we make the following key observations and reflections from a novel aesthetic perspective: 1) There is a close connection between the aesthetic attributes of an image and human perception of image quality~\cite{tinio2011image}. Specifically, the higher the conformity of an image to aesthetic standards, the higher its perceived quality; conversely, the greater the deviation from aesthetic standards, the lower the perceived quality. 2) Human aesthetic perception is not one-dimensional but hierarchical, encompassing both the perception of global harmony and the perception of local details. These two levels together form a comprehensive judgment of image aesthetics~\cite{palmer2013visual,lu2015rating}. 3) The findings in Fig.~\ref{fig:obs} demonstrate that aligning with human aesthetic preferences can effectively improve the quality of generated images and facial details~\cite{gallego2022personalizing,wu2024vmix}. This naturally raises the question: \textit{\textbf{If we adopt an anti-aesthetic alignment approach, can it achieve the goal of reducing the quality of generated images while removing facial identity?}}

Inspired by the above observations and reflections, we propose the Hierarchical Anti-Aesthetics (HAA) framework. This framework is designed to degrade image quality by hierarchically exploring aesthetic cues from global to local scales, thereby safeguarding users' facial privacy and copyright. Specifically, HAA includes the following two key branches: 1) Global Anti-Aesthetics, which weakens the overall aesthetics of images generated by malicious fine-tuners and reduces overall generation quality by constructing a global anti-aesthetic reward mechanism and designing a global anti-aesthetic loss; and 2) Local Anti-Aesthetics, which guides adversarial noise to force customized diffusion models to perform local anti-facial aesthetic alignment by establishing a local anti-aesthetic reward mechanism and designing a local anti-aesthetic loss, thereby reducing their ability to reconstruct facial details. By seamlessly integrating these branches, we form a joint hierarchical anti-aesthetics framework that fully explores aesthetic cues from global to local, thereby achieving the goal of anti-aesthetics. Implementing anti-aesthetics substantially reduces the generation quality of customized DMs, thereby reducing facial identity leakage and strengthening the protection of personal privacy and copyright. Our main contributions are as follows:

\begin{itemize}
\item From a novel aesthetic perspective, we propose a joint Hierarchical Anti-Aesthetics (HAA) framework that seamlessly integrates global and local anti-aesthetic branches. This integration effectively reduces aesthetic quality at both global and local levels, thereby achieving the goal of anti-aesthetics.

\item We introduce Global Anti-Aesthetics by constructing a global anti-aesthetic reward mechanism and designing a global anti-aesthetic loss, thereby reducing the overall generation quality of customized generative models.

\item We propose Local Anti-Aesthetics, construct a local anti-aesthetic reward mechanism, and design a local anti-aesthetic loss to guide malicious customized diffusion models to oppose alignment with human facial aesthetics.

\item Extensive experiments across multiple datasets and generative models show that our approach outperforms existing methods by a large margin, supporting the effectiveness and generalization capability of our method.
\end{itemize}

The remainder of this paper is organized as follows. Section II reviews related work on customized diffusion models and image cloaking-based privacy protection. Section III introduces the problem definition and DreamBooth preliminaries. Section IV details the proposed HAA framework, including Global and Local Anti-Aesthetics branches. Section V presents extensive experimental results and analysis. Finally, Section VI concludes the paper.

\section{RELATED WORK}
\subsection{Customized Diffusion Models}
Recent years have witnessed the continued development of diffusion models~\cite{ho2020denoising,song2020denoising,bar2022text2live,kim2022diffusionclip,ho2022cascaded,lugmayr2022repaint}, and these models have shown great potential in guiding the generation process through text input. Exemplar models, including DALL-E 2~\cite{ramesh2022hierarchical}, Stable Diffusion~\cite{rombach2022high}, and Stable Diffusion XL~\cite{podell2023sdxl}, have received widespread attention for their high sample quality. Using the latent diffusion method and incorporating CLIP-based~\cite{radford2021learning} text encoders, these models employ a two-stage process: encoding the input image or text into a latent representation and denoising within this compact latent space to bridge the gap between textual descriptions and visual content, thereby significantly improving generation quality while maintaining high fidelity and consistency in text-to-image (T2I) generation tasks.

DreamBooth~\cite{ruiz2023DreamBooth} further addresses the challenge of customized generation, thereby leading to significant improvements in generating personalized content from limited data while preserving diversity and flexibility. However, DreamBooth faces challenges with more complex customizations, particularly when personalization requires a diverse and extensive dataset. In contrast, Custom Diffusion~\cite{kumari2023multi} can first fine-tune each concept model individually and then merge them into one through constrained optimization, which enhances the ability to generate diverse outputs while maintaining coherence across different concepts. SVDiff~\cite{han2023svdiff} optimizes all singular values of the weight matrix and leverages spectral shifts to introduce a small number of trainable parameters into diffusion models, enabling them to effectively capture variations in specific themes. Textual Inversion~\cite{gal2022image} teaches a text model a new word using example images and trains its embedding to align with the corresponding visual representation by adding a new token to the vocabulary and optimizing the embedding using representative images.



\begin{figure*}[t]
  \centering
  \includegraphics[width=\textwidth]{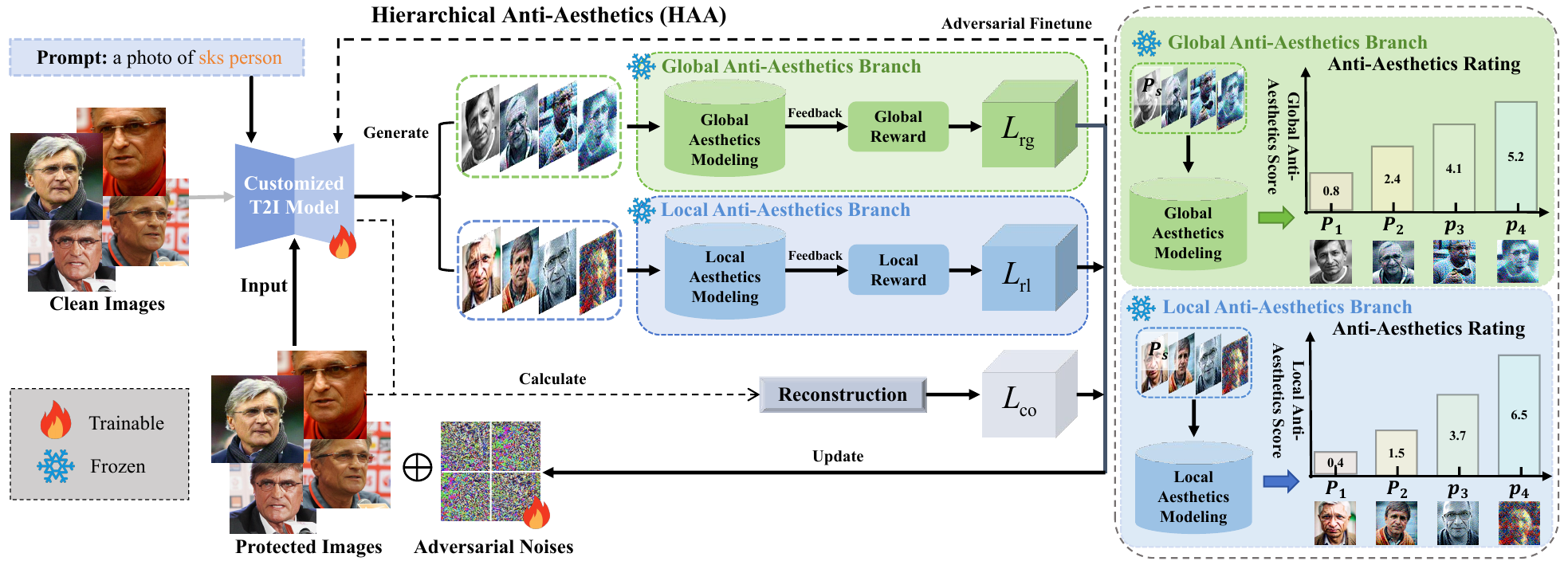}
  \caption{The framework of our proposed HAA. It is an iterative training process where the adversarial noise and the parameters of the surrogate customized generative model are updated simultaneously. The final adversarial noise generated is added to the user's image, making it resistant to malicious use by the customized generative model.}
  \label{fig:framework}
  \vspace{-10pt}
\end{figure*}

\subsection{Privacy Protection with Image Cloaking}

Recent breakthroughs in generative models, represented by diffusion models, have achieved paradigm-shifting advancements. Building on this foundation, the development of customization technologies has greatly facilitated personal visual creation. However, the misuse of personalized custom generation has raised significant concerns about copyright protection, political stability, and personal privacy. To address these issues, image cloaking methods have been developed, which involve adding specific perturbations to the original images to prevent misuse by customized generative models.

AdvDM~\cite{liang2023adversarial} performs Monte Carlo sampling on latent variables in the diffusion model's hidden space and generates perturbations specifically targeting and misleading the model's feature extraction process during the sampling time steps, ultimately leading the model to produce erroneous outputs or reducing its overall performance. \newtext{ASPL~\cite{van2023anti} employs an Alternating Surrogate and Perturbation Learning (ASPL) strategy to enhance counterattacks against the DreamBooth method}. Mist~\cite{zheng2023understanding} is specifically designed for copyright protection of artwork and can effectively defend against malicious use. CAAT~\cite{xu2024perturbing} effectively disrupts the text-to-image mapping by introducing subtle perturbations in the cross-attention layers of customized diffusion models, thereby protecting users' portrait rights from infringement. SimAC~\cite{wang2024simac} explores the limitations of the internal properties of diffusion models, investigates the relationship between time step selection and frequency domain perception as well as the roles of hierarchical features in the denoising process, and proposes an adaptive greedy time step search and a feature-based optimization framework, which enhances the anti-interference effect. \newtext{GoodAC~\cite{xu2025harnessing} further introduces a global-local anti-customization strategy by disrupting global perceptual feature correlations and local facial attributes.}

\newtext{Existing methods mainly enhance privacy protection via model-internal objectives, including reconstruction loss, cross-attention disruption, timestep/frequency manipulation, and feature- or attribute-level discrepancy. These objectives may incidentally degrade visual quality, but they do not explicitly introduce human-preference reward feedback or decompose aesthetic guidance into global image-level and local face-level cues. This underexplored use of aesthetic feedback may limit facial de-identification. To complement this line of research, HAA introduces a hierarchical anti-aesthetics perspective that explores aesthetic cues from global appearance to local facial regions, thereby improving the suppression of facial identity information and strengthening facial privacy protection.}

\section{Preliminaries}



\subsection{Problem Definition}
We study facial privacy protection against personalized diffusion models.
Under a black-box assumption, an attacker can collect a small number of the user’s publicly available face images and personalize (fine-tune) a diffusion model to generate forged portraits resembling the victim, enabling deepfakes and privacy abuse.
\newtext{Our goal is to add a magnitude-bounded, nearly imperceptible perturbation $\delta$ to each image to be shared (i.e., $x_{adv}=\mathrm{clip}(x+\delta)$ with $\|\delta\|_\infty\le\eta$), such that any model fine-tuned on these protected images produces degraded outputs (i.e., worse generation quality implies better protection) with reduced identity consistency, thereby reducing the utility of forgeries and protecting privacy.}

\subsection{DreamBooth}
DreamBooth is a classic fine-tuning technique for text-to-image diffusion models that enables personalized generation. Specifically, given a small number of subject images (3--5), it fine-tunes a pre-trained diffusion model to generate new images of that subject in various contexts. DreamBooth binds the target subject to a rare token identifier (e.g., \textit{sks}), and the input prompt follows the format of ``a \textit{sks} [class noun]'', where [class noun] represents the category (e.g., person). DreamBooth combines two objective terms: a personalization reconstruction loss and a prior preservation loss. The personalization reconstruction loss encourages the model to reconstruct the subject identity from the provided reference images, while the prior preservation loss mitigates overfitting and language drift in the few-shot setting. The overall objective is formulated as:
\begin{equation}
\begin{aligned}
\mathcal{L}_{co}(\theta)
&=\mathbb{E}_{v_0,\,c,\,t,\,\epsilon}\!\left[\left\|\epsilon-\epsilon_{\theta}\!\left(v_t,\,t,\,c\right)\right\|_2^2\right] \\
&\quad + \gamma\cdot 
\mathbb{E}_{v_0',\,c_{\text{pr}},\,t',\,\epsilon'}\!\left[\left\|\epsilon'-\epsilon_{\theta}\!\left(v'_{t'},\,t',\,c_{\text{pr}}\right)\right\|_2^2\right],
\label{eq:train}
\end{aligned}
\end{equation}
where $\epsilon_{\theta}(\cdot)$ is the noise prediction network parameterized by $\theta$ (e.g., the U-Net in latent diffusion models). $v_0$ and $v_0'$ denote the clean latents of a subject image and a class (prior) image, respectively. $c$ is the subject prompt containing the rare identifier token, and $c_{\text{pr}}$ is the class prior prompt. $t$ and $t'$ are diffusion timesteps sampled from a predefined schedule, and $\epsilon,\epsilon' \sim \mathcal{N}(0,I)$ are i.i.d.\ Gaussian noise variables. The noisy latents are constructed as:
\begin{equation}
v_t = \alpha_t v_0 + \sigma_t \epsilon,\quad
v'_{t'} = \alpha_{t'} v_0' + \sigma_{t'} \epsilon',
\end{equation}
with $(\alpha_t,\sigma_t)$ determined by the noise scheduler. $\gamma$ balances the prior preservation term. $\|\cdot\|_2$ denotes the Euclidean norm (and $\|\cdot\|_2^2$ its squared form).

\section{THE PROPOSED METHOD}

\subsection{Global Anti-Aesthetics (GAA)}
We start from a novel perspective—the relationship between aesthetics and quality—and introduce a novel Global Anti-Aesthetics Algorithm (GAA), which focuses on mining global aesthetic cues to degrade the overall quality of images. Specifically, we construct a global anti-aesthetic reward mechanism and design a global anti-aesthetic loss function in conjunction with a reconstruction loss to train adversarial noise.

\textbf{Global anti-aesthetic reward mechanism.} Given a sample $x$, we perform diffusion steps on it using a pre-trained surrogate DM. We then employ a Monte Carlo sampling strategy to sample the diffusion timestep (and the corresponding noise) and conduct a \emph{single-step} prediction (i.e., one forward pass of the denoiser followed by VAE decoding) for global anti-aesthetic alignment. This process can be formalized as follows:
\begin{equation}
x_t' = G_{\theta} (N(E(x^{adv}_t),B(T)), C),
\label{eq:sample}
\end{equation}
where \( x^{adv}_t \) denotes the adversarial sample at the current optimization step (we use \(t\) to match the notation in Eq.~\ref{eq:sample}), \( B(\cdot) \) denotes a Monte Carlo sampler, and \( T \) stands for the set of diffusion timesteps. Concretely, \(B(T)\) samples one timestep \(t\in T\) (and the associated Gaussian noise used by the scheduler) for each image in the mini-batch, yielding a noisy latent through the noise scheduler \(N(\cdot)\) after VAE encoding \(E(\cdot)\). \newtext{Equivalently, for image \(i\), \(v_i^{adv}=E(x_i^{adv})\) and \(v_{t_i}^{adv}=\alpha_{t_i}v_i^{adv}+\sigma_{t_i}\epsilon_i\) before \(G_\theta(\cdot)\).} Then, \(G_{\theta}(\cdot)\) performs a direct one-step prediction and decodes the predicted latent back to the image domain.
This pipeline is efficient and end-to-end differentiable with respect to \(x^{adv}\) because it only consists of differentiable components (VAE encoder/decoder, noise scheduler, and the denoiser forward pass), without running the full reverse diffusion chain (see Eq.~15 in~\cite{ho2020denoising}).

To implement GAA, we construct a global anti-aesthetic reward model to obtain global anti-aesthetic rewards. For this purpose, we utilize the publicly available human aesthetic preference dataset VisionRewardDB-Image~\cite{xu2024imagereward} to train a Global Aesthetic Reward Model \(RM_g\) with BLIP as the backbone. Then, \(RM_g\) is frozen. Through \(RM_g\), we can conduct a global aesthetic evaluation of the generated images to obtain global aesthetic scores. Subsequently, we construct global anti-aesthetic rewards $R_{i}^{g}$ based on these global aesthetic scores. The formulas can be expressed as follows:
\begin{equation}
r_i^g = - RM_g(x_{ti}', C_g),
\label{eq:r}
\end{equation}
\begin{equation}
R_{i}^{g}=\frac{r_{i}^{g}-\frac{1}{n}\sum_{i = 1}^{n}r_{i}^{g}}{\sqrt{\frac{1}{n}\sum_{i = 1}^{n}(r_{i}^{g}-\frac{1}{n}\sum_{i = 1}^{n}r_{i}^{g})^{2}}},
\label{eq:rig}
\end{equation}
Let \( x_{ti}' \) denote the \( i \)-th generated sample in the batch at the \( t \)-th iteration, where \( n \) denotes the batch size. \( C_g \) represents the global prompt. Eq.~\ref{eq:r} is used to calculate the global anti-aesthetic reward, aiming to measure deviations in aesthetics. This mechanism guides the model to perform ``reverse optimization'' for global aesthetics. Eq.~\ref{eq:rig} serves to normalize the global anti-aesthetic reward. Through normalization, we adjust the reward value to an appropriate range and avoid excessively large or small numerical values. The larger the value of $R_{i}^{g}$, the higher the global anti-aesthetic score and the poorer the overall quality of the generated image.

\textbf{Global anti-aesthetic loss.} In our approach, we utilize the global anti-aesthetic reward $R_{i}^{g}$ obtained from Eq.~\ref{eq:rig} to calculate the global anti-aesthetic loss $\mathcal{L}_{rg}$. This loss is then employed to update the adversarial noise, thereby undermining the overall generation quality of malicious fine-tuners. The specific formula is as follows:
\begin{equation}
L_{rg}=-\frac{1}{n}\sum_{i=1}^{n}
\log\!\Big(1+\exp\!\big(-R^{g}_{i}\big)\Big),
\label{eq:6}
\end{equation}
\begin{equation}
x_{adv}=Proj(\arg\max_{x_{adv}}(\mathcal{L}_{co}+\lambda \cdot \mathcal{L}_{rg})),
\end{equation}
where \( \exp(\cdot) \) denotes the exponential function, \(\mathcal{L}_{co}\) is the reconstruction loss defined in Eq.~\ref{eq:train}, and \(\lambda\) is a balancing coefficient. $Proj(\cdot)$ ensures that the adversarial noise remains within a prescribed $\ell_{\infty}$ budget. \newtext{Specifically, the projection is applied to the perturbation $\delta$, while clipping is only used to map $x+\delta$ back to the valid image range.} Inspired by reinforcement learning, where the optimization process is differentiable~\cite{yu2022surprising}, we compute $\mathcal{L}_{rg}$ based on the rewards $R_i^{g}$ provided by $RM_g$. Specifically, Eq.~\ref{eq:6} applies a softplus function to smoothly map the global anti-aesthetic rewards $R_i^{g}$ into an averaged, differentiable objective. Maximizing this objective encourages larger $R_i^{g}$, thereby promoting misalignment with human aesthetic preferences and guiding adversarial perturbations to degrade the generation quality of the T2I model.
In practice, gradients are back-propagated from \(RM_g(\cdot)\) to \(x^{adv}\) through \(x_t'\) in Eq.~\ref{eq:sample}, enabling direct optimization of the perturbation under the \(\ell_\infty\) constraint.

Inspired by adversarial learning, we adopt an adversarial game strategy. On one hand, we train a surrogate diffusion model (SDM) with \( x_{adv} \) generated by GAA to enhance robustness. On the other hand, we attack the robust SDM to generate stronger attacks. The training process of the SDM can be represented as follows:
\begin{equation}
\theta'= \arg\min_{\theta'} \mathcal{L}_{co}(x_{adv}, \theta'),
\label{eq:8}
\end{equation}
where \(\theta'\) denotes the parameters of the SDM. Through this adversarial mechanism, adversarial samples with stronger attack effects can be generated, thereby improving the overall attack capability. This strategy is also employed in the subsequent methods.

\subsection{Local Anti-Aesthetics (LAA)}
Human perception of aesthetics is multidimensional, encompassing not only global aesthetic perception but also local aesthetic perception. Inspired by this, we propose a novel Local Anti-Aesthetics Algorithm (LAA), where we construct a local anti-aesthetic reward mechanism and design a local anti-aesthetic loss. This approach forces maliciously fine-tuned DMs to perform anti-facial aesthetic alignment, thereby enhancing the ability to eliminate facial identity cues.

\textbf{Local anti-aesthetic reward mechanism.} Specifically, a lightweight face detection model $F$ is applied to the $x'_{t}$ defined in Eq.~\ref{eq:sample} to obtain face bounding boxes $Box_t$ and confidence scores $S$. Similar to the global aesthetic reward model, to mine local facial aesthetic cues, we use the same BLIP architecture to construct a low-cost local facial aesthetic reward model \(RM_l\). Using the detected boxes and the generated images, we extract the corresponding facial regions and use \(RM_l\) to provide local aesthetic feedback. We then utilize the feedback to calculate the local anti-aesthetic rewards $R^l_i$ for updating the adversarial noise. The lower the quality of the generated face, the higher the $R^l_i$. Let \( C_l \) represent the local prompt; this process can be formalized as follows:
\begin{equation}
S_{ti}, Box_{ti} = F(x_{ti}'),
\label{eq:face}
\end{equation}
\begin{equation}
r_i^l = - RM_l(x_{ti}',Box_{ti}, C_l),
\label{eq:10}
\end{equation}
\begin{equation}
R_{i}^{l}=\frac{r_{i}^{l}-\frac{1}{n}\sum_{i = 1}^{n}r_{i}^{l}}{\sqrt{\frac{1}{n}\sum_{i = 1}^{n}(r_{i}^{l}-\frac{1}{n}\sum_{i = 1}^{n}r_{i}^{l})^{2}}}.
\label{eq:11}
\end{equation}

\textbf{Local anti-aesthetic loss.} Based on $R^l_i$, we further calculate the reward loss \(\mathcal{L}_{rl}\) to update the adversarial noise. This process can be formalized as follows:
\begin{equation}
L_{rl}=-\frac{1}{n}\sum_{i=1}^{n}
\log\!\Big(1+\exp\!\big(-R^{l}_{i}\big)\Big),
\label{eq:12}
\end{equation}
\begin{equation}
x_{adv}=Proj(\arg\max_{x_{adv}}(\mathcal{L}_{co}+\beta \cdot \mathcal{L}_{rl})),
\label{eq:13}
\end{equation}
where \(\beta\) is a balancing coefficient. Similarly, the parameters of the SDM are updated during the iterative attack. By attacking a more robust surrogate diffusion model, the attack strength is enhanced, thereby achieving the local anti-aesthetic objective.

\begin{algorithm}[t]
\caption{Hierarchical Anti-Aesthetics (HAA)}
\label{alg:haa}
\begin{algorithmic}[1]
\Require Clean images $\mathcal{X}$; subject prompt $C$; global prompt $C_g$; local prompt $C_l$; timestep set $\mathcal{T}$;
batch size $n$; iterations $K$; step size $\alpha$; weights $\lambda,\beta$.
\Require Frozen reward models $RM_g, RM_l$; face detector $F$; trainable surrogate diffusion model $SDM_{\theta'}$.
\Ensure Protected images \newtext{$\mathcal{X}_{adv}=\{\mathrm{clip}(x+\delta) \mid x\in\mathcal{X}\}$} with $\|\delta\|_\infty\le\eta$.

\State Initialize perturbation $\delta \gets 0$
\State Sample a mini-batch $\{x_i\}_{i=1}^{n} \sim \mathcal{X}$
\For{$k=1$ to $K$}
    \State Sample timesteps $\{t_i\}_{i=1}^{n}\sim \mathcal{T}$ via $B(\mathcal{T})$
    \For{$i=1$ to $n$}
        \State $x^{adv}_i \gets \mathrm{clip}(x_i+\delta)$
        \State $x'_{t_i} \gets G_{\theta}\!\big(N(E(x^{adv}_i), t_i), C\big)$ \Comment{Eq.~\ref{eq:sample}}
        \State $r^g_i \gets -RM_g(x'_{t_i}, C_g)$ \Comment{Eq.~\ref{eq:r}}
        \State $(S_i,Box_i) \gets F(x'_{t_i})$ \Comment{Eq.~\ref{eq:face}}
        \State $r^l_i \gets -RM_l(x'_{t_i}, Box_i, C_l)$ \Comment{Eq.~\ref{eq:10}}
    \EndFor
    \State $R^g \gets \mathrm{Normalize}(r^g)$ \Comment{Eq.~\ref{eq:rig}}
    \State $R^l \gets \mathrm{Normalize}(r^l)$ \Comment{Eq.~\ref{eq:11}}
    \State $L_{rg} \gets -\frac{1}{n}\sum_{i=1}^{n}\log\!\big(1+\exp(-R^g_i)\big)$ \Comment{Eq.~\ref{eq:6}}
    \State $L_{rl} \gets -\frac{1}{n}\sum_{i=1}^{n}\log\!\big(1+\exp(-R^l_i)\big)$ \Comment{Eq.~\ref{eq:12}}
    \State $L_{co} \gets \mathcal{L}_{DreamBooth}(SDM_{\theta'}, \{x^{adv}_i\}, C)$ \Comment{Eq.~\ref{eq:train}}
    \State $L_{\mathrm{total}} \gets L_{co} + \lambda L_{rg} + \beta L_{rl}$ \Comment{Eq.~\ref{eq:total_loss}}
    \State \newtext{$\delta \gets \Pi_{\|\delta\|_\infty\le\eta}\Big(\delta + \alpha\cdot\mathrm{sign}(\nabla_{\delta}L_{\mathrm{total}})\Big)$}
    \Comment{Eq.~\ref{eq:15}}
    \State $\theta' \gets \theta' - \mu\cdot\nabla_{\theta'} L_{co}$ \Comment{Eq.~\ref{eq:8}}
\EndFor
\State \Return $\mathcal{X}_{adv}$
\end{algorithmic}
\end{algorithm}

\begin{table}[!t]
\centering
\caption{Performance on CelebA-HQ Dataset.}
\label{tab:cele}
\resizebox{0.46\textwidth}{!}{
\begin{tabular}{@{}l|cccc@{}}
\toprule
\multicolumn{5}{c}{CelebA-HQ} \\
\toprule
Method & \multicolumn{4}{c}{``a photo of sks person''} \\
\cmidrule{2-5}
& FDSR↓ & Face Similarity↓ & Image Reward↓ & FID↑ \\
\midrule
Clean & 1.000 & 0.498 & 0.599 & 112.3 \\
Mist & 0.969 & 0.375 & 0.191 & 281.2 \\
CAAT & 0.813 & 0.339 & 0.326 & 271.7 \\
ASPL & 0.750 & 0.332 & 0.006 & 359.5 \\
SimAC & 0.438 & 0.321 & -0.188 & 388.9 \\
\newtext{GoodAC} & \newtext{0.375} & \newtext{0.192} & \newtext{-0.278} & \newtext{421.1} \\
HAA & \textbf{0.281} & \textbf{0.117} & \textbf{-0.894} & \textbf{471.1} \\
\midrule
Method & \multicolumn{4}{c}{``a dslr portrait of sks person''} \\
\cmidrule{2-5}
& FDSR↓ & Face Similarity↓ & Image Reward↓ & FID↑ \\
\midrule
Clean & 0.859 & 0.346 & 0.706 & 175.1 \\
Mist & 0.859 & 0.308 & 0.249 & 271.0 \\
CAAT & 0.766 & 0.260 & 0.236 & 251.8 \\
ASPL & 0.656 & 0.251 & -0.183 & 384.0 \\
SimAC & 0.500 & 0.239 & -0.202 & 369.0 \\
\newtext{GoodAC} & \newtext{0.375} & \newtext{0.196} & \newtext{-0.575} & \newtext{435.5} \\
HAA & \textbf{0.266} & \textbf{0.115} & \textbf{-0.861} & \textbf{461.4} \\
\midrule
Method & \multicolumn{4}{c}{``a close-up photo of sks person, high details''} \\
\cmidrule{2-5}
& FDSR↓ & Face Similarity↓ & Image Reward↓ & FID↑ \\
\midrule
Clean & 0.625 & 0.276 & 0.281 & 200.8 \\
Mist & 0.594 & 0.268 & -0.128 & 268.4 \\
CAAT & 0.510 & 0.177 & -0.419 & 305.9 \\
ASPL & 0.438 & 0.167 & -0.882 & 433.3 \\
SimAC & 0.333 & 0.162 & -0.779 & 401.7 \\
\newtext{GoodAC} & \newtext{0.219} & \newtext{0.112} & \newtext{-1.276} & \newtext{462.1} \\
HAA & \textbf{0.177} & \textbf{0.077} & \textbf{-1.334} & \textbf{476.0} \\
\midrule
Method & \multicolumn{4}{c}{``a photo of sks person looking at the mirror''} \\
\cmidrule{2-5}
& FDSR↓ & Face Similarity↓ & Image Reward↓ & FID↑ \\
\midrule
Clean & 0.672 & 0.276 & 0.022 & 232.9 \\
Mist & 0.695 & 0.257 & -0.289 & 287.0 \\
CAAT & 0.602 & 0.171 & -0.545 & 316.1 \\
ASPL & 0.523 & 0.157 & -0.854 & 422.4 \\
SimAC & 0.562 & 0.156 & -0.773 & 405.6 \\
\newtext{GoodAC} & \newtext{0.531} & \newtext{0.128} & \newtext{-1.192} & \newtext{455.9} \\
HAA & \textbf{0.289} & \textbf{0.085} & \textbf{-1.296} & \textbf{473.5} \\
\bottomrule
\end{tabular}}
\vspace{-10pt}
\end{table}


\begin{table}[!t]
\centering
\caption{Performance on VGGFace2 Dataset.}
\label{tab:vgg}
\resizebox{0.46\textwidth}{!}{
\begin{tabular}{@{}l|cccc@{}}
\toprule
\multicolumn{5}{c}{VGGFace2} \\
\toprule
Method & \multicolumn{4}{c}{``a photo of sks person''} \\
\cmidrule{2-5}
& FDSR↓ & Face Similarity↓ & Image Reward↓ & FID↑ \\
\midrule
Clean & 0.875 & 0.421 & 0.625 & 199.1 \\
Mist & 0.906 & 0.227 & 0.248 & 355.6 \\
CAAT & 0.719 & 0.145 & 0.371 & 349.8 \\
ASPL & 0.750 & 0.193 & 0.373 & 388.6 \\
SimAC & 0.500 & 0.118 & 0.143 & 435.6 \\
\newtext{GoodAC} & \newtext{0.375} & \newtext{0.111} & \newtext{-0.070} & \newtext{442.4} \\
HAA & \textbf{0.125} & \textbf{0.031} & \textbf{-0.252} & \textbf{468.6} \\

\midrule
Method & \multicolumn{4}{c}{``a dslr portrait of sks person''} \\
\cmidrule{2-5}
& FDSR↓ & Face Similarity↓ & Image Reward↓ & FID↑ \\
\midrule
Clean & 0.813 & 0.379 & 0.717 & 227.6 \\
Mist & 0.813 & 0.240 & 0.409 & 357.2 \\
CAAT & 0.750 & 0.179 & 0.559 & 332.8 \\
ASPL & 0.703 & 0.173 & 0.351 & 387.9 \\
SimAC & 0.422 & 0.102 & -0.018 & 429.3 \\
\newtext{GoodAC} & \newtext{0.344} & \newtext{0.085} & \newtext{-0.105} & \newtext{439.4} \\
HAA & \textbf{0.109} & \textbf{0.028} & \textbf{-0.608} & \textbf{462.4} \\
\midrule
Method & \multicolumn{4}{c}{``a close-up photo of sks person, high details''} \\
\cmidrule{2-5}
& FDSR↓ & Face Similarity↓ & Image Reward↓ & FID↑ \\
\midrule
Clean & 0.688 & 0.376 & 0.720 & 232.3 \\
Mist & 0.573 & 0.215 & 0.279 & 343.2 \\
CAAT & 0.521 & 0.155 & 0.353 & 341.1 \\
ASPL & 0.490 & 0.150 & 0.039 & 381.0 \\
SimAC & 0.292 & 0.078 & -0.581 & 434.5 \\
\newtext{GoodAC} & \newtext{0.219} & \newtext{0.054} & \newtext{-0.944} & \newtext{440.0} \\
HAA & \textbf{0.083} & \textbf{0.019} & \textbf{-1.162} & \textbf{458.4} \\
\midrule
Method & \multicolumn{4}{c}{``a photo of sks person looking at the mirror''} \\
\cmidrule{2-5}
& FDSR↓ & Face Similarity↓ & Image Reward↓ & FID↑ \\
\midrule
Clean & 0.750 & 0.355 & 0.482 & 246.0 \\
Mist & 0.641 & 0.188 & 0.005 & 354.9 \\
CAAT & 0.633 & 0.139 & 0.145 & 334.0 \\
ASPL & 0.602 & 0.153 & -0.055 & 380.6 \\
SimAC & 0.414 & 0.092 & -0.616 & 429.0 \\
\newtext{GoodAC} & \newtext{0.406} & \newtext{0.090} & \newtext{-0.927} & \newtext{434.8} \\
HAA & \textbf{0.258} & \textbf{0.031} & \textbf{-1.119} & \textbf{446.0} \\
\bottomrule
\end{tabular}}
\vspace{-10pt}
\end{table}

\subsection{Hierarchical Anti-Aesthetics (HAA)}
\newtext{Building upon the complementary strengths of GAA and LAA, we propose a unified HAA framework (Fig.~\ref{fig:framework}) that integrates global and local aesthetic degradation within a single optimization objective. This design disrupts image generation quality at multiple perceptual levels, achieving anti-aesthetic effects in both global composition and local facial details. Specifically, the hierarchy in HAA is realized through the reward-model inputs and the joint objective, rather than a separate sampling schedule: GAA evaluates the full decoded image with the global prompt \(C_g\), while LAA evaluates detected facial regions with the local prompt \(C_l\). Compared with methods relying solely on \(\mathcal{L}_{co}\) or feature-level discrepancy objectives, HAA further incorporates two frozen reward-guided terms, \(\mathcal{L}_{rg}\) and \(\mathcal{L}_{rl}\), while using the same timestep sampler to maintain efficient differentiable feedback.} Accordingly, the final training objective is defined as:

\newtext{HAA uses the standard projected adversarial-perturbation loop as the basis for controlled comparison. The technical change is the optimization signal: HAA injects frozen human-preference feedback into both whole-image and face-local perturbation updates. These terms are not generic auxiliary losses; they define external preference targets that penalize preference-aligned realism and facial detail quality. Thus, the contribution lies in reward-guided hierarchical anti-aesthetic supervision, rather than in a new outer projected-gradient optimizer.}

\begin{equation}
\mathcal{L}_{total} = \mathcal{L}_{co}+\lambda \cdot \mathcal{L}_{rg}+\beta \cdot \mathcal{L}_{rl}.
\label{eq:total_loss}
\end{equation}
\begin{equation}
x_{adv}=Proj(\arg\max_{x_{adv}}(\mathcal{L}_{total})).
\label{eq:15}
\end{equation}

In each iteration, the overall loss is maximized to optimize the adversarial samples, enabling the adversarial noise to poison malicious fine-tuners and disrupt the global and local aesthetics of the generated images, thereby degrading overall generation quality. While the adversarial samples are iteratively optimized, the parameters of the SDM are fine-tuned in an adversarial training-like manner. Ultimately, through iterative attacks, the final adversarial samples can effectively remove facial identity information, thereby protecting facial privacy and copyright.

\section{EXPERIMENTS}
\subsection{Experimental Settings}
\noindent \textbf{Datasets.} 
Following previous work~\cite{xu2024perturbing}, we conduct experiments on the CelebA-HQ~\cite{zhu2022celebv} and VGGFace2~\cite{cao2018vggface2} datasets. Both are large, classic datasets specifically for facial privacy research. Due to the wide range of comparison metrics, we select 10 individuals from each dataset, covering different genders, ages, and ethnicities, with at least 15 images for each individual. Unless explicitly stated, we conduct experimental tests uniformly on CelebA-HQ.

\noindent \textbf{Models.}
Our experiments primarily utilize the popular open-source diffusion model SD-v2.1~\cite{rombach2021highresolution}. Furthermore, to extensively validate the effectiveness of our approach, we also conduct experiments on SD-v1.4 and SD-v1.5. In addition, extended experiments on mainstream SD-V3.0~\cite{esser2024scaling} and FLUX~\cite{labs2025flux1kontextflowmatching} are also conducted in the subsequent experiments.

\noindent \textbf{Baselines and comparisons.}
We compare our method with various state-of-the-art methods designed to prevent the misuse of user images by DMs, including ASPL~\cite{van2023anti}, Mist~\cite{zheng2023understanding}, CAAT~\cite{xu2024perturbing}, SimAC~\cite{wang2024simac}, and \newtext{GoodAC~\cite{xu2025harnessing}}.

\noindent \textbf{Metrics.}
Following previous work~\cite{xu2024perturbing}, we employ a classic face detection model to detect the generated images and calculate Facial Detection Success Rate (FDSR)~\cite{yang2015facial}. For identity consistency, we calculate Face Similarity~\cite{deng2019arcface} between the detected faces and clean images. \newtext{FDSR is the proportion of generated images with at least one detected face. Face Similarity is the average cosine similarity between \(\ell_2\)-normalized ArcFace embeddings; if no face is detected, the per-image similarity is set to 0. Moreover, we use Image Reward~\cite{xu2024imagereward} to assess the quality of the generated images, and report its score as IR.} Finally, Fréchet Inception Distance (FID)~\cite{heusel2017gans} is used to assess the similarity between the generated images and clean images. Lower FDSR and Face Similarity indicate weaker face detectability and identity consistency, while lower Image Reward and higher FID indicate poorer preference-aligned generation quality; together, these metrics indicate stronger privacy protection.

\noindent \textbf{Implementation Details.}
To ensure a fair comparison, we adopt a unified experimental setup. We use the latest Stable Diffusion (v2.1) as the pre-trained model and fine-tune the text encoder and UNet models using the DreamBooth method for 1000 training steps with a batch size of 4 and a learning rate of $5 \times 10^{-7}$. We set $\alpha=5\times 10^{-3}$ as the step size, along with a perturbation budget $\eta=0.05$. Moreover, in the LAA module, we employ RetinaFace~\cite{deng2020retinaface} for face detection because of its high efficiency and accuracy. \newtext{For computing Face Similarity, we use ArcFace~\cite{deng2019arcface} for its high accuracy and robustness.} All experiments in the table are run three times, and the results are averaged to ensure stability.

\newtext{Our evaluation is designed to test both effectiveness and generalization. In addition to the main CelebA-HQ and VGGFace2 comparisons under four prompts, we include component ablations, prompt mismatch, varying numbers of protected images, perturbation-budget studies, image-transformation and purification defenses, transfer across Stable Diffusion versions, transfer to SD-V3.0 and FLUX, and three fine-tuning pipelines (SVDiff, Textual Inversion, and Custom Diffusion).}

\begin{figure}[t]
 \centering
 \includegraphics[width=0.8\linewidth]{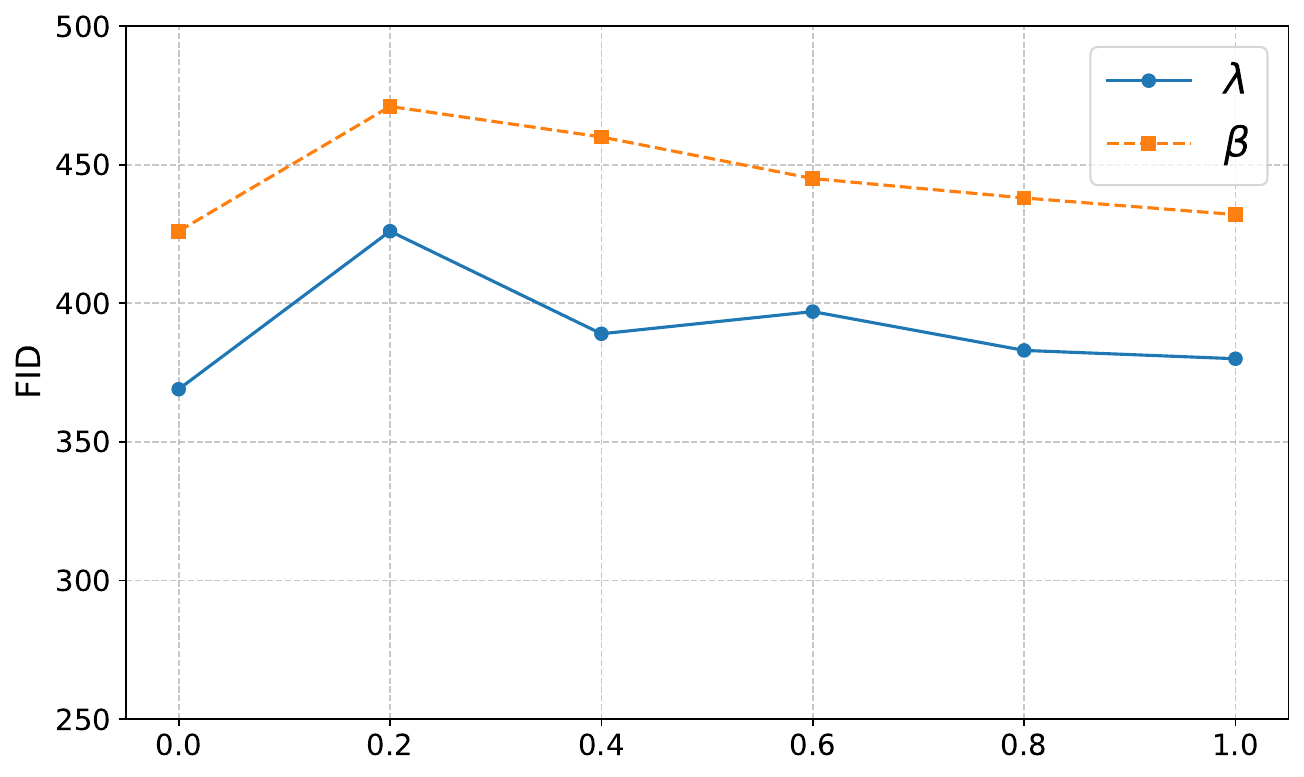}
 \caption{FID value varies with different $\lambda$ and $\beta$ values defined in Eq.~\ref{eq:total_loss} on CelebA-HQ.}
 \label{fig:para}
 \vspace{-10pt}
\end{figure}
\begin{figure}[t]
 \centering
 \includegraphics[width=0.95\linewidth]{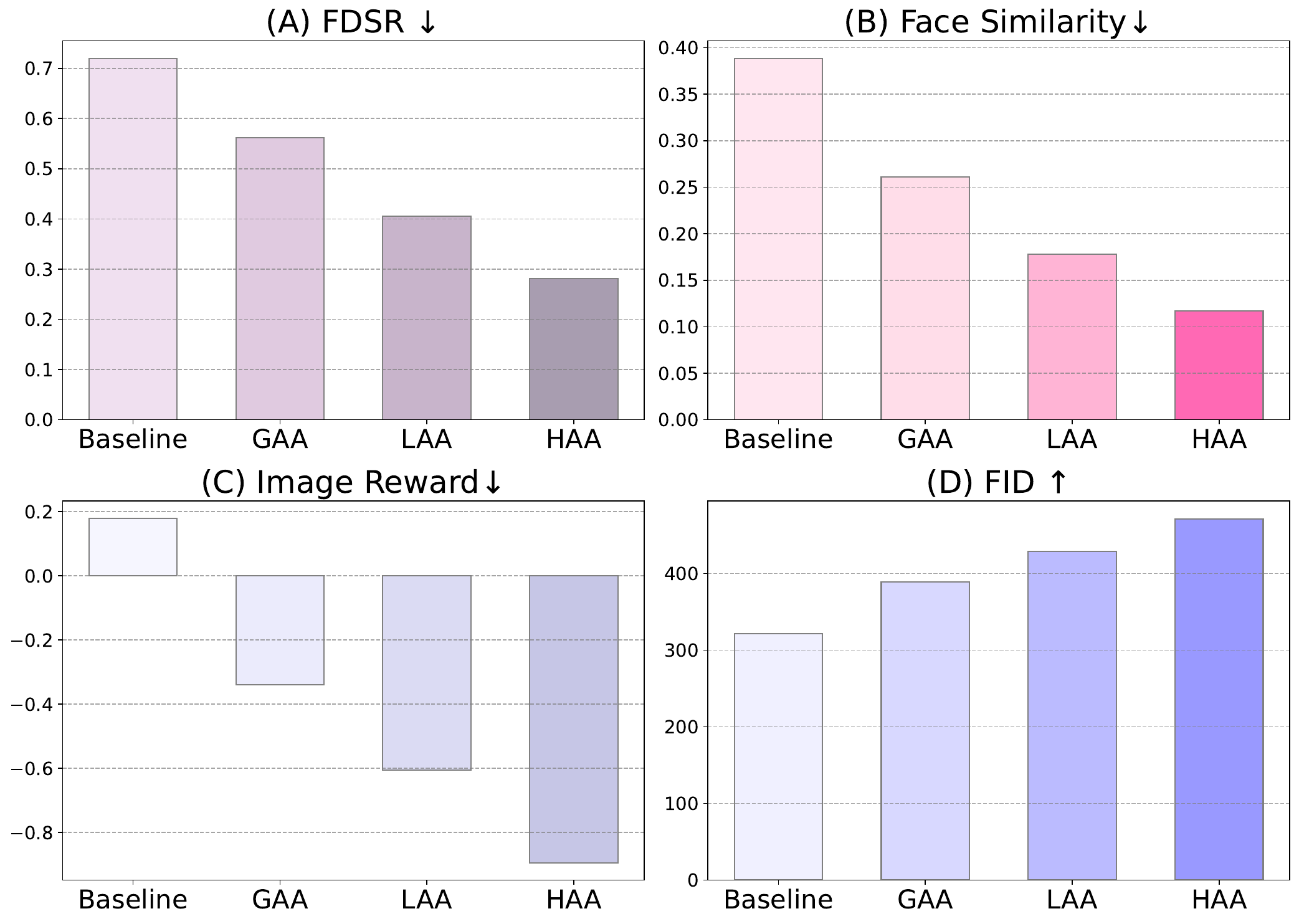}
 \caption{Effectiveness Analysis of HAA Components on CelebA-HQ.}
 \label{fig:abl}
 \vspace{-10pt}
\end{figure}

\subsection{Comparison with Baselines}
To more thoroughly assess effectiveness, we perform quantitative comparisons on two datasets (CelebA-HQ and VGGFace2) under four prompt settings, three of which remain unseen during training, and report results against recent state-of-the-art methods in Tab.~\ref{tab:cele} and Tab.~\ref{tab:vgg}. For each prompt, we randomly generate 32 images and report the mean of four metrics. On CelebA-HQ with the prompt ``a photo of sks person'', HAA reduces FDSR from 96.9\% to 28.1\% and Face Similarity from 37.5\% to 11.7\%, lowers Image Reward from 0.191 to -0.894, and increases FID from 281.2 to 471.1 relative to Mist. On VGGFace2 with the prompt ``a close-up photo of sks person, high details'', HAA reduces FDSR from 52.1\% to 8.3\% and Face Similarity from 15.5\% to 1.9\%, lowers Image Reward from 0.353 to -1.162, and increases FID from 341.1 to 458.4 relative to CAAT.

\begin{table}[!t]
\centering
\caption{Assessment across different DM versions on CelebA-HQ.}
\label{tab:diff}
\resizebox{0.46\textwidth}{!}{
\begin{tabular}{c|c|c|c|c}
\toprule
\multicolumn{5}{c}{SD-V1.4} \\
\hline
attack &  FDSR↓ & Face Similarity↓ & Image Reward↓ & FID↑ \\
\hline
Clean           & 0.938       & 0.439       & 0.440       & 148.4     \\
Mist            & 0.906       & 0.332       & -0.027      & 237.8     \\
CAAT            & 0.125       & 0.035       & -1.083      & 508.8     \\
ASPL            & 0.375       & 0.076       & -0.728      & 458.4     \\
SimAC           & 0.186       & 0.067       & -1.079      & 508.1   \\
HAA & \textbf{0.062} & \textbf{0.030} & \textbf{-1.577} & \textbf{540.4}\\
\toprule
\multicolumn{5}{c}{SD-V1.5} \\
\hline
attack &  FDSR↓ & Face Similarity↓ & Image Reward↓ & FID↑ \\
\hline
Clean & 0.969 & 0.454 & 0.577 & 140.4 \\
Mist & 1.000 & 0.383 & -0.165 & 225.9 \\
CAAT & 0.312 & 0.036 & -1.335 & 466.5 \\
ASPL & 0.250 & 0.059 & -0.904 & 479.4 \\
SimAC & 0.219 & 0.149 & -1.230 & 452.8 \\
HAA & \textbf{0.000} & \textbf{0.003} & \textbf{-1.563} & \textbf{523.7} \\
\toprule
\end{tabular}}
\vspace{-15pt}
\end{table}
\begin{table}[!t]
\centering
\caption{Effectiveness studies across different fine-tuning scenarios on CelebA-HQ.}
\label{tab:fine-tuning}
\resizebox{0.46\textwidth}{!}{
\begin{tabular}{@{}l|cccc@{}}
\toprule
\multicolumn{5}{c}{Different fine-tuning scenarios} \\
\toprule
Method & \multicolumn{4}{c}{SVDiff} \\
\cmidrule{2-5}
& FDSR↓ & Face Similarity↓ & Image Reward↓ & FID↑ \\
\midrule
Clean           & 0.914       & 0.335       & 0.538       & 149.2       \\ 
Mist            & 0.875       & 0.202       & 0.274       & 212.9       \\ 
CAAT            & 0.688       & 0.199       & 0.121       & 337.4       \\ 
ASPL            & 0.516       & 0.168       & 0.103       & 388.3       \\ 
SimAC           & 0.789       & 0.205       & 0.095       & 289.6       \\ 
HAA   & \textbf{0.492} & \textbf{0.133} & \textbf{0.002} & \textbf{401.2} \\
\midrule
Method & \multicolumn{4}{c}{Textual Inversion} \\
\cmidrule{2-5}
& FDSR↓ & Face Similarity↓ & Image Reward↓ & FID↑ \\
\midrule
Clean           & 0.969       & 0.258       & 0.914       & 140.8       \\ 
Mist            & 1.000       & 0.212       & 0.432       & 185.7       \\ 
CAAT            & 0.969       & \textbf{0.172}       & 0.543       & 191.5       \\ 
ASPL            & 1.000       & 0.178       & 0.535       & 152.9       \\ 
SimAC           & 1.000       & 0.234       & 0.415       & 148.1       \\ 
HAA   & \textbf{0.938} & 0.194 & \textbf{0.030} & \textbf{213.5} \\ 
\midrule
Method & \multicolumn{4}{c}{Custom Diffusion} \\
\cmidrule{2-5}
& FDSR↓ & Face Similarity↓ & Image Reward↓ & FID↑ \\
\midrule
Clean          & 0.938       & 0.437       & 0.567       & 120.0       \\ 
Mist            & 0.906       & 0.330       & 0.173       & 194.8       \\ 
CAAT             & 0.875       & 0.308       & 0.177       & 271.3       \\ 
ASPL             & 0.938       & 0.319       & 0.315       & 271.2       \\ 
SimAC            & 0.969       & 0.364       & 0.480       & 202.0       \\ 
HAA   & \textbf{0.781} & \textbf{0.254} & \textbf{0.152} & \textbf{275.4} \\
\midrule

\end{tabular}}
\vspace{-15pt}
\end{table}

Across datasets and prompt variations, these results suggest that HAA more effectively suppresses identity-related cues in the generated outputs, as indicated by the concurrent decreases in FDSR and Face Similarity. We also observe that stronger identity removal coincides with degraded perceptual quality (higher FID) and lower preference-aligned scores (lower Image Reward), which is consistent with the goal of discouraging high-quality personalized reconstruction under malicious fine-tuning. We attribute this behavior to the hierarchical anti-aesthetics design: the global branch broadly disrupts aesthetics that correlate with overall realism, while the local branch further targets facial regions where identity information concentrates, and their combination yields a more consistent de-identification effect under prompts not seen during training.

\begin{table}[!t]
\centering
\caption{Prompt mismatch between training and testing on CelebA-HQ. The training prompt is ``a photo of sks person'', while the inference prompt uses rare identifiers such as ``sks'' or ``t@t''.}
\label{tab:mismatch}
\begin{tabular}{c|c|cccc}
\toprule
\multicolumn{6}{c}{``a photo of [v] person''} \\
\midrule
Train [v] & Test [v] & FDSR $\downarrow$ & FS $\downarrow$ & IR $\downarrow$ & FID $\uparrow$ \\
\midrule
sks & sks & \textbf{0.281} & \textbf{0.117} & \textbf{-0.894} & \textbf{471.1} \\
\midrule
sks & t@t & 0.438 & 0.316 & 0.017 & 344.6 \\
\midrule
\multicolumn{6}{c}{``a dslr portrait of [v] person''} \\
\midrule
Train [v] & Test [v] & FDSR $\downarrow$ & FS $\downarrow$ & IR $\downarrow$ & FID $\uparrow$ \\
\midrule
sks & sks & \textbf{0.266} & \textbf{0.115} & \textbf{-0.861} & \textbf{461.4} \\
sks & t@t & 0.578 & 0.221 & 0.197 & 276.5 \\
\midrule
\multicolumn{6}{c}{``a close-up photo of [v] person, high details''} \\
\midrule
Train [v] & Test [v] & FDSR $\downarrow$ & FS $\downarrow$ & IR $\downarrow$ & FID $\uparrow$ \\
\midrule
sks & sks & \textbf{0.177} & \textbf{0.077} & \textbf{-1.334} & \textbf{476.0} \\
sks & t@t & 0.385 & 0.148 & -0.610 & 325.1 \\
\midrule
\multicolumn{6}{c}{``a photo of [v] person looking at the mirror''} \\
\midrule
Train [v] & Test [v] & FDSR $\downarrow$ & FS $\downarrow$ & IR $\downarrow$ & FID $\uparrow$ \\
\midrule
sks & sks & \textbf{0.289} & \textbf{0.085} & \textbf{-1.296} & \textbf{473.5} \\
sks & st@t & 0.523 & 0.138 & -0.678 & 322.0 \\
\bottomrule
\end{tabular}
\vspace{-10pt}
\end{table}

\subsection{Parameter Tuning}
In our formulation, Eq.~\ref{eq:total_loss} introduces two hyperparameters, \(\lambda\) and \(\beta\), which weight the global and local anti-aesthetic losses, respectively. We perform parameter tuning on CelebA-HQ using SD-2.1 and examine how these weights affect FID.

As shown in Fig.~\ref{fig:para}, \(\lambda\) exerts a pronounced influence on FID. When \(\lambda\) increases from small values, FID rises accordingly, which is consistent with the global anti-aesthetic objective increasingly dominating optimization and thereby pushing generated samples farther from the clean distribution. However, beyond \(\lambda=0.2\), FID begins to decrease, suggesting diminishing returns---and potentially a change in the optimization regime in which overly strong global degradation does not further widen the distributional gap captured by FID. Based on this empirical trend, we set \(\lambda=0.2\) in the remaining experiments. We observe that \(\beta\) also materially affects overall performance. Following the same tuning protocol, we set \(\beta=0.2\), which provides a favorable operating point under our evaluation setting by balancing the contribution of local facial degradation against the other objectives in Eq.~\ref{eq:total_loss}.


\begin{table}[!t]
\centering
\caption{Effectiveness assessment by varying the number of perturbed images on CelebA-HQ.}
\begin{tabular}{c|c|cccc}
\hline
\multicolumn{6}{c}{``a photo of sks person''} \\ \hline
\textbf{Perturbed} & \textbf{Clean} & \textbf{FDSR↓} & \textbf{FS↓} & \textbf{IR↓} & \textbf{FID↑} \\ \hline
4 & 0 & \textbf{0.281} & \textbf{0.117} & \textbf{-0.894} & \textbf{471.1} \\ \hline
3 & 1 & 0.625 & 0.290 & -0.164 & 338.8 \\ \hline
2 & 2 & 0.688 & 0.390 & 0.210 & 267.3 \\ \hline
1 & 3 & 0.719 & 0.426 & 0.273 & 213.6 \\ \hline
0 & 4 & 1.000 & 0.498 & 0.599 & 112.3 \\ \hline
\multicolumn{6}{c}{``a dslr portrait of sks person''} \\ \midrule
\textbf{Perturbed} & \textbf{Clean} & \textbf{FDSR↓} & \textbf{FS↓} & \textbf{IR↓} & \textbf{FID↑} \\ \hline
4 & 0 & \textbf{0.266} & \textbf{0.115} & \textbf{-0.861} & \textbf{461.4} \\ \hline
3 & 1 & 0.641 & 0.244 & 0.055 & 320.1 \\ \hline
2 & 2 & 0.703 & 0.310 & 0.262 & 269.4 \\ \hline
1 & 3 & 0.688 & 0.330 & 0.230 & 215.1 \\ \hline
0 & 4 & 0.859 & 0.346 & 0.706 & 175.1 \\ \hline
\multicolumn{6}{c}{``a close-up photo of sks person, high details''} \\ \hline
\textbf{Perturbed} & \textbf{Clean} & \textbf{FDSR↓} & \textbf{FS↓} & \textbf{IR↓} & \textbf{FID↑} \\ \hline
4 & 0 & \textbf{0.177} & \textbf{0.077} & \textbf{-1.334} & \textbf{476.0} \\ \hline
3 & 1 & 0.427 & 0.164 & -0.663 & 373.5 \\ \hline
2 & 2 & 0.479 & 0.223 & -0.357 & 319.0 \\ \hline
1 & 3 & 0.458 & 0.262 & -0.138 & 238.8 \\ \hline
0 & 4 & 0.625 & 0.276 & 0.281 & 200.8 \\ \hline
\multicolumn{6}{c}{``a photo of sks person looking at the mirror''} \\ \hline
\textbf{Perturbed} & \textbf{Clean} & \textbf{FDSR↓} & \textbf{FS↓} & \textbf{IR↓} & \textbf{FID↑} \\ \hline
4 & 0 & \textbf{0.289} & \textbf{0.085} & \textbf{-1.296} & \textbf{473.5} \\ \hline
3 & 1 & 0.547 & 0.164 & -0.668 & 378.0 \\ \hline
2 & 2 & 0.602 & 0.227 & -0.404 & 320.7 \\ \hline
1 & 3 & 0.594 & 0.274 & -0.256 & 265.9 \\ \hline
0 & 4 & 0.672 & 0.276 & 0.022 & 232.9 \\ \hline
\end{tabular}
\label{tab:perturbation}
\vspace{-10pt}
\end{table}

\subsection{Ablation Study}
Effectiveness Analysis of HAA Components. We conduct systematic ablation studies to examine the individual roles of Global Anti-Aesthetics (GAA) and Local Anti-Aesthetics (LAA), as well as their combined effect. As reported in Fig.~\ref{fig:abl}, relative to the Baseline (FDSR \(=0.719\)), adding GAA reduces FDSR to \(0.562\) (a \(21.8\%\) relative reduction) and decreases Face Similarity to \(0.261\) (a \(32.7\%\) relative reduction). This behavior is consistent with GAA primarily acting at the image-level distribution: by discouraging globally preferred aesthetic configurations, it tends to suppress holistic realism cues that also correlate with identity consistency.

When we instead incorporate the local anti-facial aesthetics loss (LAA) into the Baseline, LAA yields stronger privacy-oriented degradation than GAA in this setting: FDSR further drops to \(0.406\) (a \(27.8\%\) relative reduction), and Image Reward decreases from \(-0.339\) to \(-0.605\). These results suggest that explicitly targeting facial regions provides a more direct lever for weakening identity-relevant evidence, because many recognition pipelines depend on localized facial details that may persist even when global image quality degrades.

Finally, the full HAA model, which combines both global and local anti-aesthetic objectives, shows the most pronounced effect: FDSR decreases to \(0.281\) (a \(60.9\%\) relative reduction vs.\ the Baseline), Face Similarity decreases to \(0.117\) (a \(69.8\%\) relative reduction), Image Reward reaches \(-0.894\), and FID increases to \(471.1\) (a \(46.5\%\) relative increase). The stepwise trend across variants indicates that GAA and LAA are complementary rather than redundant: the global branch broadly shifts samples away from high-fidelity, preference-aligned generations, while the local branch concentrates the disruption on facial regions where identity information is densest. Their combination yields a more consistent degradation of both detection success and identity similarity, which aligns with the goal of reducing malicious fine-tuners' ability to reconstruct recognizable faces under our experimental setting.

\subsection{\newtext{Necessity of the Global Branch}}
\newtext{Although LAA directly targets detected faces, identity leakage in customized diffusion models is not restricted to the detected face crop. Hair, head contour, pose, clothing, lighting, and scene co-occurrence can still support subject-token binding and identity reconstruction. In addition, LAA depends on reliable face detection. GAA therefore provides a principled full-image anti-aesthetic objective for complementary image-level suppression and remains effective when local detection is unreliable.}

\begin{figure}[!t]
\centering
\includegraphics[width=0.8\linewidth]{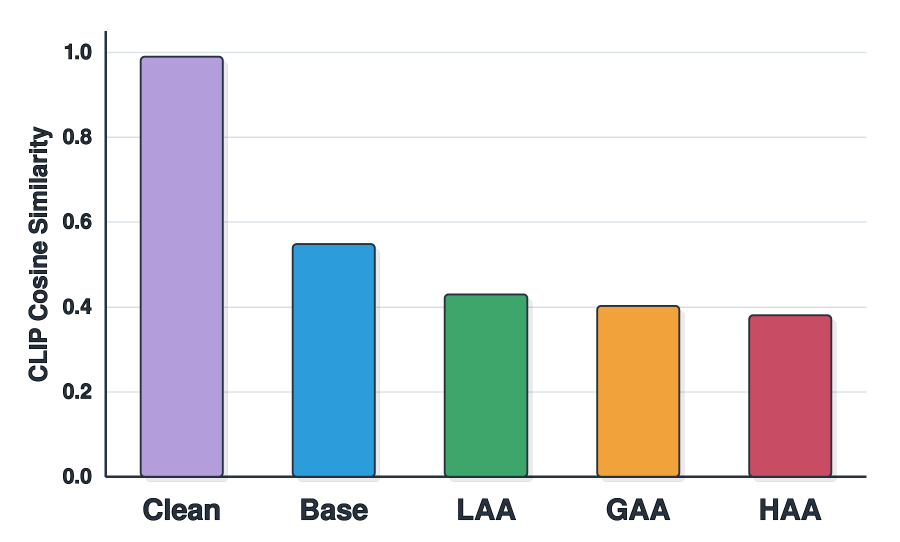}
\caption{\newtext{CLIP cosine similarity after masking detected face regions. Lower values indicate less residual non-face identity-related information.}}
\label{fig:r2q3_clipcos}
\vspace{-10pt}
\end{figure}

\begin{figure}[!t]
\centering
\includegraphics[width=\linewidth]{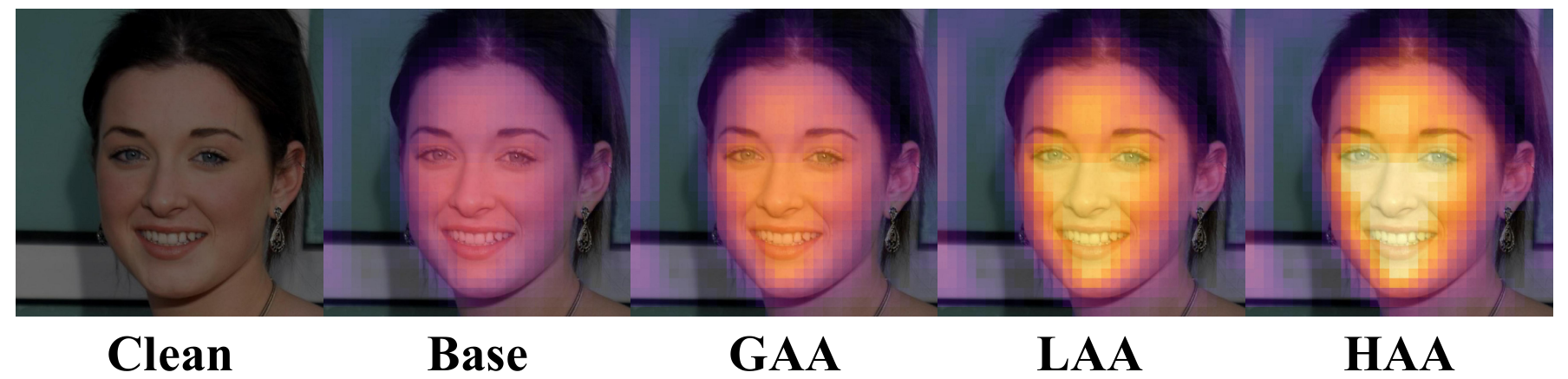}
\caption{\newtext{Grad-CAM response visualization for HAA components. Brighter regions indicate stronger optimization pressure for suppressing identity-supporting information.}}
\label{fig:r2q3_cooperate}
\vspace{-10pt}
\end{figure}

\newtext{We first mask all detected face regions and compute CLIP image-embedding cosine similarity between the masked generated images and clean references. Clean denotes unprotected training, and Base denotes reconstruction-loss-only perturbation without anti-aesthetic losses. As shown in Fig.~\ref{fig:r2q3_clipcos}, GAA reduces residual non-face similarity more than LAA (0.403 vs. 0.430), and HAA achieves the lowest value (0.381). This indicates that identity-supporting information also exists outside the face crop and that the global branch helps remove it.} \newtext{The quantitative ablation in Fig.~\ref{fig:abl} further shows that GAA alone improves over the baseline, LAA provides stronger face-local degradation, and HAA achieves the best overall protection. Fig.~\ref{fig:r2q3_cooperate} visualizes the corresponding Grad-CAM response regions, with response strength following Clean \(<\) Base \(<\) GAA \(<\) LAA \(<\) HAA. LAA focuses on facial details, GAA covers broader image-level structures, and HAA yields the broadest response.}

\begin{figure}[!t]
\centering
\includegraphics[width=\linewidth]{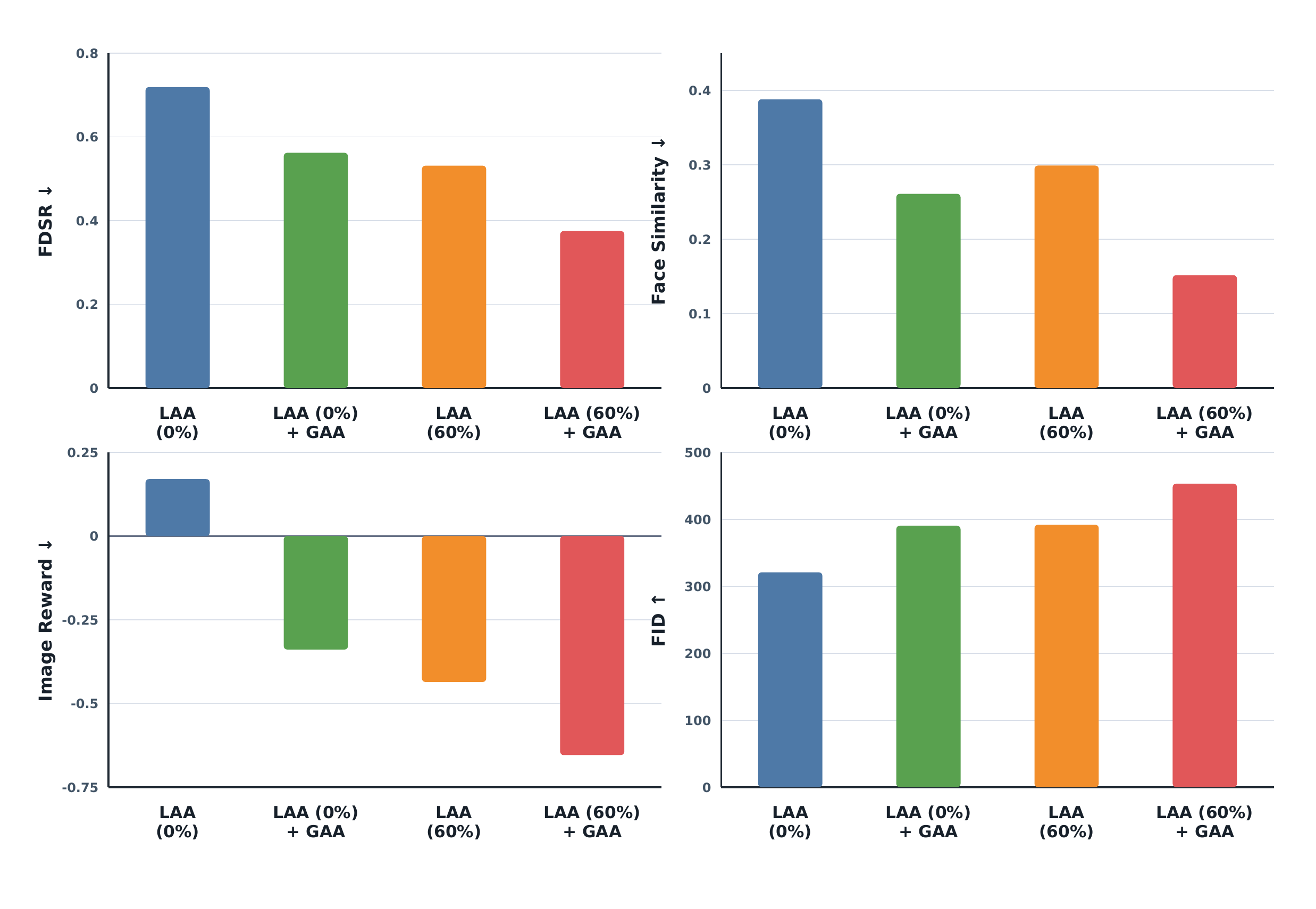}
\caption{\newtext{Face-detection-failure stress test. The 0\% and 60\% settings denote valid-local-detection availability for LAA. Adding GAA consistently improves LAA when local detections are unavailable or partially available.}}
\label{fig:r2q3_failure}
\vspace{-8pt}
\end{figure}

\newtext{Finally, we evaluate a face-detection-failure stress setting. In Fig.~\ref{fig:r2q3_failure}, LAA (0\%) corresponds to the case where no valid face box is available, so the local reward cannot provide effective face-local guidance. Adding GAA improves LAA at both 0\% and 60\% valid-local-detection availability. When 60\% of local detections are available, adding GAA reduces FDSR from 0.531 to 0.375 and Face Similarity from 0.299 to 0.152, while increasing FID from 391.9 to 453.2. These results indicate that GAA provides complementary feedback when the local branch is weakened, rather than merely increasing perturbation strength.}

\subsection{Black-Box Performance on Different DM Versions.} 
To better approximate practical deployments where the target customized T2I model differs from the surrogate used for perturbation crafting, we evaluate HAA under black-box model-version mismatch by transferring attacks across different Stable Diffusion releases. As shown in Tab.~\ref{tab:diff}, HAA maintains strong effectiveness on both SD-v1.4 and SD-v1.5. Concretely, on SD-v1.5, it further reduces FDSR to \(0.000\) and Face Similarity to \(0.003\), with Image Reward \(-1.563\) and FID \(523.7\). Compared to baselines, these results indicate that perturbations transfer across model versions and suppress both face detectability and identity consistency. We view this stability as evidence that HAA primarily disrupts version-invariant cues shared across model releases, rather than fragile version-specific artifacts. In particular, degrading global appearance statistics while targeting facial regions reduces downstream fine-tuning's reliance on semantically meaningful identity features. This helps preserve protection strength when the attacker's model differs from the one used during perturbation optimization.

\begin{table}[!t]
\centering
\caption{Effectiveness studies across varying noise budgets on CelebA-HQ.}
\label{tab:eps}
\resizebox{0.46\textwidth}{!}{
\begin{tabular}{c|c|c|c|c}
\hline
$\eta$ & FDSR $\downarrow$ & Face Similarity $\downarrow$ & Image Reward $\downarrow$ & FID $\uparrow$ \\
\hline
0.00 &  1.000 & 0.498 & 0.599 & 112.3 \\
0.02 & 0.425 & 0.220 & -0.428 & 379.8 \\
0.05 & 0.281 & 0.117 & -0.894 & 471.1 \\
0.10 & 0.156 & 0.054 & -0.999 & 462.2 \\
\textbf{0.15} & \textbf{0.000} & \textbf{0.000} & \textbf{-1.364} & \textbf{497.0} \\
\hline
\end{tabular}}
\vspace{-10pt}
\end{table}

\begin{table}[!t]
\centering
\caption{Assessment across SD-V3.0 and FLUX under a black-box setting on CelebA-HQ.}
\label{tab:sd3}
\resizebox{0.95\linewidth}{!}{
\begin{tabular}{c|c|c|c|c}
\hline
\textbf{Model} & \textbf{Attack} & \textbf{Face Similarity↓} & \textbf{Image Reward↓} & \textbf{FID↑} \\ \hline
SD-V3.0 & CAAT & 0.357 & 0.725 & 170.4 \\
SD-V3.0 & SimAC & 0.353 & 0.988 & 188.5 \\
SD-V3.0 & HAA & 0.343 & 0.561 & 197.0 \\
FLUX & CAAT & 0.399 & 0.630 & 123.1 \\
FLUX & SimAC & 0.397 & 0.453 & 117.7 \\
FLUX & HAA & 0.375 & 0.336 & 148.5 \\ \hline
\end{tabular}}
\vspace{-10pt}
\end{table}

\subsection{More fine-tuning scenarios}
Tab.~\ref{tab:fine-tuning} evaluates HAA's transferability across three customization pipelines (SVDiff, Textual Inversion, Custom Diffusion) on CelebA-HQ. Overall, HAA consistently shifts fine-tuned generators to weaker identity evidence (lower Face Similarity/FDSR) and reduced preference-aligned quality (lower Image Reward), supporting perturbation effectiveness beyond the DreamBooth-style surrogate used in optimization.

In the SVDiff setting, HAA outperforms other methods, achieving the best protection metrics: lowest FDSR (\(0.492\)), Face Similarity (\(0.133\)), Image Reward (\(0.002\)), and highest FID (\(401.2\)), indicating a stronger ability to disrupt face detection and identity consistency. Tab.~\ref{tab:fine-tuning} supports HAA's generalization across heterogeneous fine-tuning strategies. Its impact is most pronounced in pipelines directly affecting facial detail generation (SVDiff, Custom Diffusion), while Textual Inversion is intrinsically harder to defend and CAAT obtains the lowest Face Similarity in that setting. Even so, HAA reliably degrades preference-aligned quality and increases distributional divergence, fulfilling the goal of reducing maliciously customized models' utility for privacy-invasive face reconstruction.


\begin{table}[h]
\centering
\caption{\newtext{HAA under diverse image transformation and purification defenses with $\eta=0.02$ on CelebA-HQ.}}
\label{tab:def}
\resizebox{0.46\textwidth}{!}{
\large
\begin{tabular}{c|c|c|c|c}
\hline
$Method $ & FDSR $\downarrow$ & Face Similarity $\downarrow$ & Image Reward $\downarrow$ & FID $\uparrow$ \\
\hline
Clean & 1.000 & 0.498 & 0.599 & 112.3 \\
HAA & \textbf{0.425} & \textbf{0.220} & \textbf{-0.428 } & \textbf{379.8} \\
Gaussian blur k=3 & 0.644 & 0.302 & 0.315 & 215.1 \\
Gaussian blur k=5 & 0.875 & 0.326 & 0.377 & 215.6 \\
Gaussian blur k=7 & 0.906 & 0.349 & 0.410 &209.1\\
JPEG q=30 & 0.938 & 0.269 & -0.337 & 257.2 \\
JPEG q=50 & 0.869 & 0.262 & -0.288 & 263.2 \\
JPEG q=70 & 0.813 & 0.254 &-0.173 &267.0\\
\newtext{DiffPure} & \newtext{0.906} & \newtext{0.375} & \newtext{0.451} & \newtext{235.8} \\
\hline
\end{tabular}
    }
\vspace{-10pt}
\end{table}

\subsection{Prompt mismatch assessment}
Tab.~\ref{tab:mismatch} evaluates robustness to prompt mismatch on CelebA-HQ. We optimize perturbations using a fixed training prompt (``a photo of sks person'') and then test at inference time with prompt templates that vary in style and, importantly, replace the identifier token with rare alternatives (e.g., ``t@t'' or ``st@t''). This setting probes whether the protection hinges on a particular token string or instead transfers to semantically equivalent prompts. Across all prompts, HAA effectively reduces identity-related information. As shown in Tab.~\ref{tab:mismatch}, when the test identifier matches training (\([v]=\)``sks''), HAA achieves strong protection, with low FDSR (e.g., \(0.281\), \(0.177\)), low Face Similarity (down to \(0.077\)), and negative Image Rewards (as low as \(-1.334\)). These results offer two complementary insights. First, HAA's effect is partly token-agnostic, as the perturbations consistently weaken identity signals. Second, the performance gap between matched and mismatched identifiers reveals that fine-tuning relies on the specific token-to-identity link; altering the token weakens this association and, thus, the protection. 


\subsection{Assessment on number of perturbed images}
Tab.~\ref{tab:perturbation} evaluates HAA's protection against the number of perturbed training images, assessing its effectiveness and sample efficiency. Across four prompts, 4 perturbed images consistently yield the strongest protection across all metrics. For example, with ``a photo of sks person'', 4 images achieve FDSR=0.281, Face Similarity=0.117, Image Reward=-0.894, and FID=471.1. Notably, HAA remains effective with just 1 perturbed image: for ``a dslr portrait of sks person'', 1 perturbed image reduces FDSR to 0.688 and Face Similarity to 0.330 vs.\ the clean baseline; for ``a close-up photo of sks person, high details'', it lowers Image Reward to -0.138 and raises FID to 238.8, indicating limited poisoning degrades image quality and identity consistency. Moreover, protection strength scales predictably with perturbed image count. For ``a photo of sks person looking at the mirror'', increasing perturbed images from 0 to 4 steadily improves metrics: FDSR drops from \(0.672\) to \(0.289\), Face Similarity from \(0.276\) to \(0.085\), and FID rises from \(232.9\) to \(473.5\), suggesting more poisoned images prevent perturbation averaging and reduce identity reconstruction reliability. HAA also performs well with detail-rich prompts: for ``a close-up photo of sks person, high details'', 2 perturbed images achieve FDSR=0.479 and Image Reward=-0.357. Overall, Tab.~\ref{tab:perturbation} supports HAA's practicality for constrained budgets and predictable protection scaling with perturbed image count.

\subsection{Effectiveness studies across varying noise budgets} We study how the perturbation budget affects the generation quality of customized models when applying HAA. Tab.~\ref{tab:eps} reports results for \(\eta \in \{0.0, 0.02, 0.05, 0.1, 0.15\}\). As \(\eta\) increases, the generated outputs degrade more noticeably (e.g., FDSR drops from \(1.000\) to \(0.000\) and Face Similarity from \(0.498\) to \(0.000\)), indicating stronger suppression of identity cues. At the same time, overly large budgets are less practical because they can visibly affect the protected images. Notably, even under a tight budget of \(\eta=0.02\), HAA remains effective (FDSR \(0.425\), Face Similarity \(0.220\), Image Reward \(-0.428\), FID \(379.8\)), supporting its applicability under realistic constraints.

\begin{figure}[t]
 \centering
 \includegraphics[width=\linewidth]{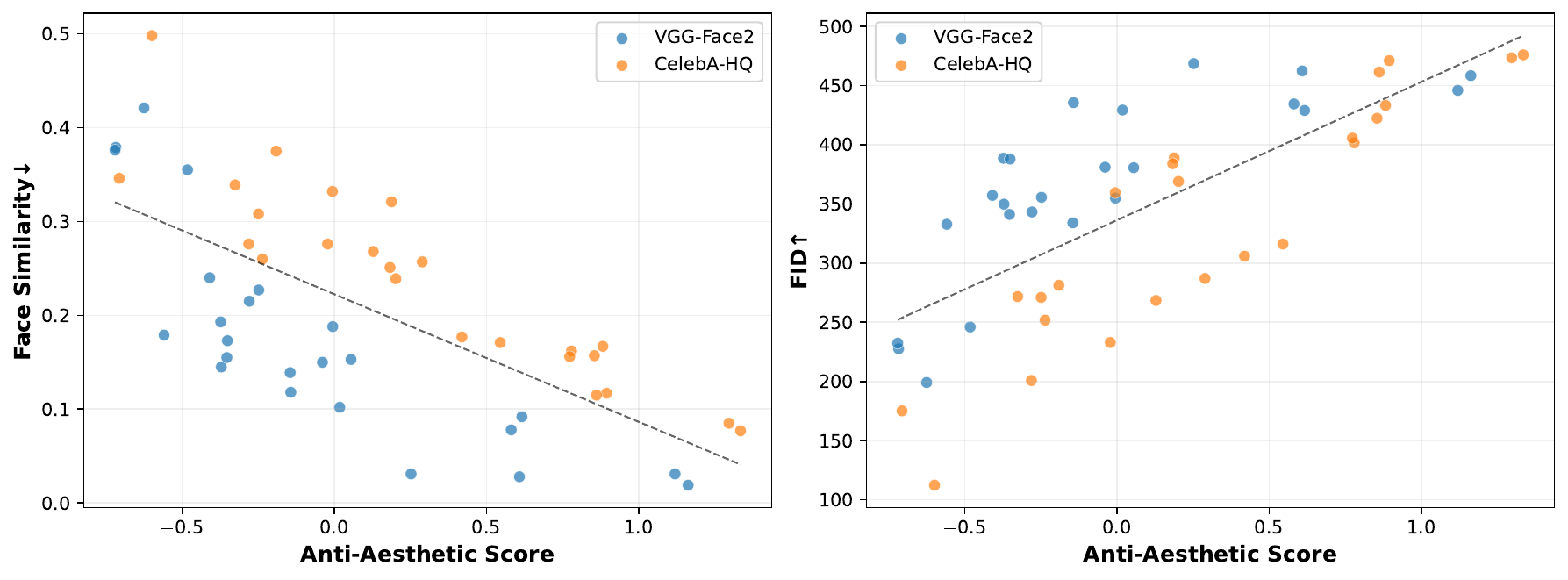}
 \caption{The relationship between Anti-Aesthetic Score, Face Similarity (reflecting identity-removal capability), and FID (reflecting image quality). The blue and orange scatter points represent the VGGFace2 and CelebA-HQ datasets, respectively.}
 \label{fig:score_relation}
 \vspace{-10pt}
\end{figure}
\begin{figure}[t]
 \centering
 \includegraphics[width=0.8\linewidth]{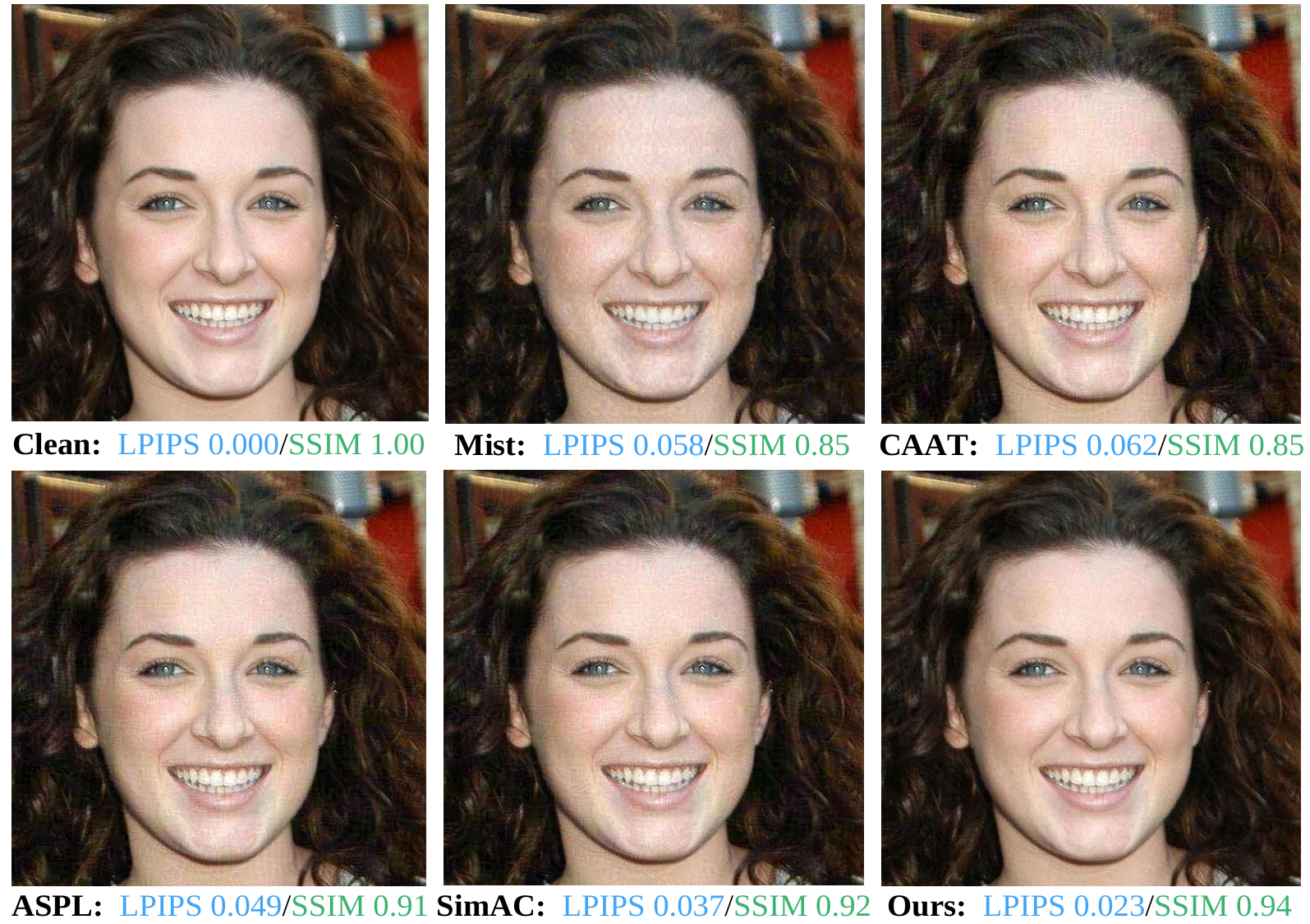}
 \caption{Portrait image protected by different methods. The protective noise added by HAA on user images is subtle on CelebA-HQ.}
 \label{fig:show}
 \vspace{-12pt}
\end{figure}
\begin{figure}[t]
  \centering
  \includegraphics[width=0.95\linewidth]{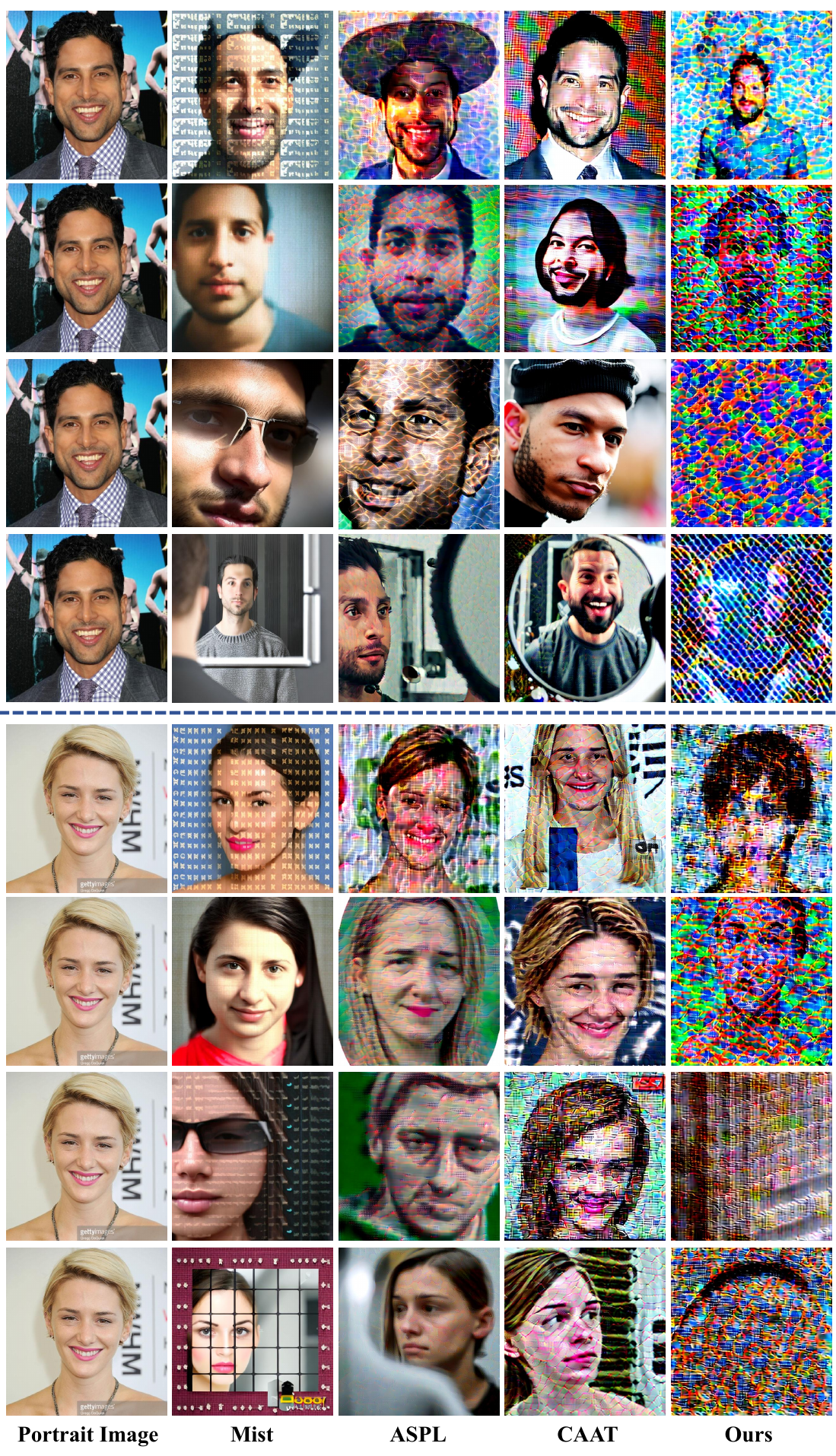}
  \caption{Comparative visualization results on the VGGFace2 dataset. Two individuals and four prompts are used for experimental evaluation. Arranged from top to bottom, the prompts for each individual are as follows: \newtext{``a photo of a sks person''}, \newtext{``a dslr portrait of a sks person''}, \newtext{``a close-up photo of a sks person with high details''}, and \newtext{``a photo of a sks person looking at the mirror''}. The poorer the image quality, the better the protection effect and the more effectively identity information is removed, demonstrating the advantage of HAA's privacy-preserving performance.}
  \label{fig:visual}
  \vspace{-10pt}
\end{figure}
\begin{figure*}[t]
 \centering
 \includegraphics[width=0.89\textwidth]{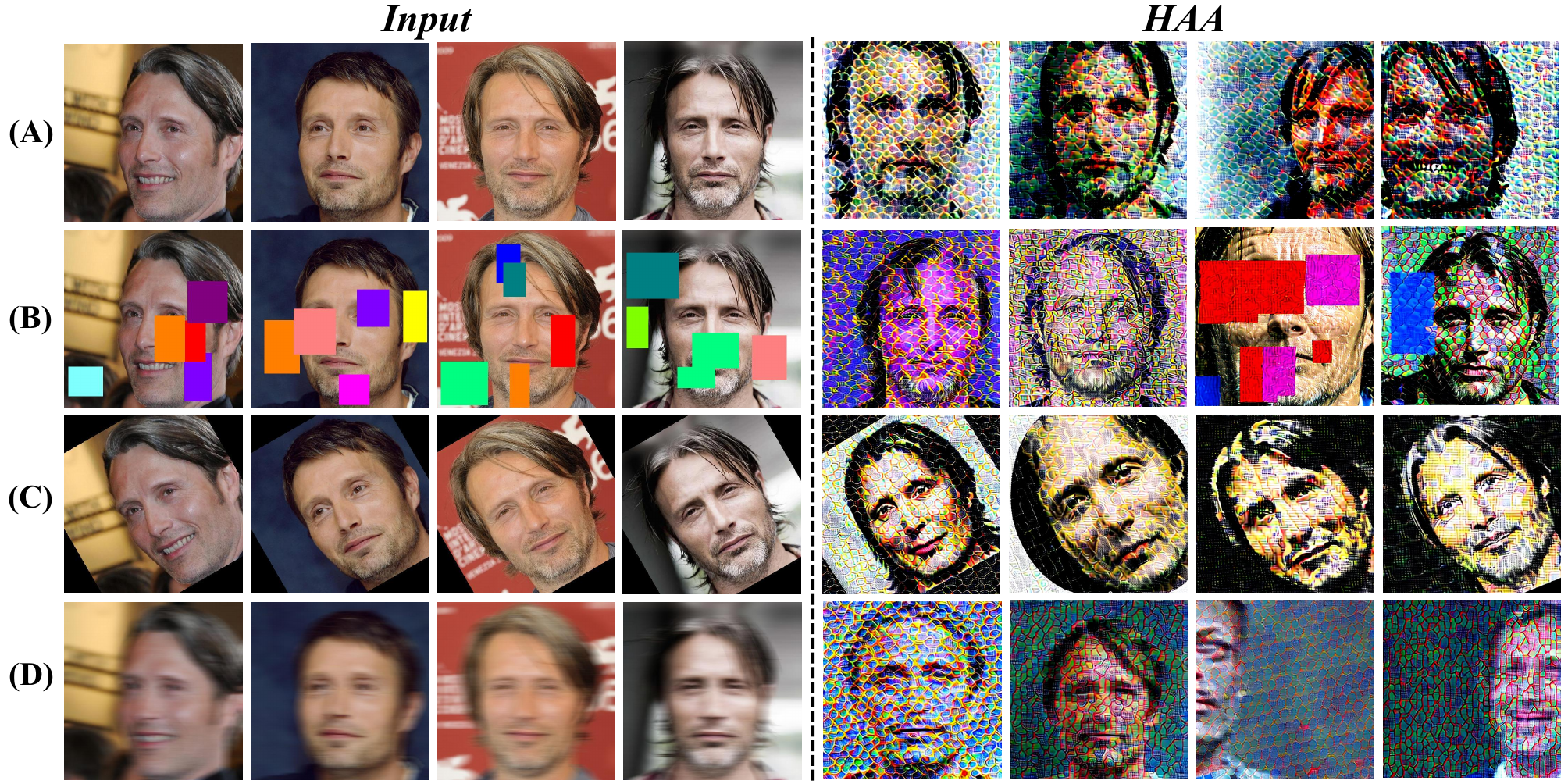}
 \caption{Visualization of HAA's protection effectiveness under extreme scenarios such as occlusion, rotation, and motion blur. (A) represents the original image, (B) represents random occlusion on the original image, (C) represents rotating the original image by 30 degrees, and (D) represents applying motion blur to the image. The visualization results show that our method remains effective in extreme scenarios.}
 \label{fig:show5}
 \vspace{-10pt}
\end{figure*}

\subsection{Assessment across SD-V3.0 and FLUX models}
Tab.~\ref{tab:sd3} evaluates black-box transfer to SD-V3.0 and FLUX on CelebA-HQ using Face Similarity, Image Reward, and FID. On SD-V3.0, HAA attains the lowest Face Similarity (\(0.343\)) and the lowest Image Reward (\(0.561\)) among the compared methods, while its FID (\(197.0\)) is higher than CAAT and SimAC. On FLUX, HAA again yields the lowest Face Similarity (\(0.375\)) and the lowest Image Reward (\(0.336\)), and it also produces the highest FID (\(148.5\)).

Overall, these results indicate that HAA remains effective under black-box transfer to recent generative models, but the absolute transfer strength is weaker than what we observe on SD-v1.4/v1.5 in our other experiments. A plausible explanation is architectural mismatch: our surrogate is built on SD-2.1 (similar to SD-v1.4/v1.5), which uses a U-Net denoiser, whereas SD-V3.0 and FLUX adopt transformer-based denoisers. This shift likely reduces gradient and feature alignment across models, thereby weakening perturbation transfer. We also observe a consistent trade-off: stronger identity suppression and lower preference-aligned quality (lower Face Similarity and Image Reward) coincide with larger distributional deviation (higher FID), suggesting that on these newer backbones, protection gains are more tightly coupled with perceptible degradation in generation quality under our evaluation protocol.

\subsection{Robustness of HAA} 
Tab.~\ref{tab:def} evaluates HAA's robustness to defensive processing with a \(\eta=0.02\) budget. Compared to clean images (FDSR \(=1.000\), Face Similarity \(=0.498\), FID \(=112.3\)), HAA degrades identity and quality signals: FDSR drops to \(0.425\), Face Similarity to \(0.220\), Image Reward becomes negative (\(-0.428\)), and FID increases to \(379.8\). We first test common non-adaptive edits, including Gaussian blur and JPEG compression. Although these operations weaken the perturbation and increase FDSR/Face Similarity, Image Reward remains lower and FID remains higher than the clean baseline, indicating that HAA retains a protective effect under common image processing.

\newtext{We further evaluate DiffPure as a purification-based post-processing defense. The experimental results indicate that DiffPure weakens HAA's protection in this setting, increasing FDSR to \(0.906\) and Face Similarity to \(0.375\). Nevertheless, it does not fully restore the generation behavior under the clean-image condition: Face Similarity remains lower than that of the clean setting (\(0.375\) vs.\ \(0.498\)), Image Reward is lower (\(0.451\) vs.\ \(0.599\)), and FID remains substantially higher (\(235.8\) vs.\ \(112.3\)). Therefore, under our experimental setting, the protective effect of HAA remains observable.}

\subsection{Correlation Verification between Anti-Aesthetic and Privacy Protection}
Fig.~\ref{fig:score_relation} reports the relationship between the anti-aesthetic score (\(r_i^g\) in Eq.~\ref{eq:r}; higher values indicate stronger aesthetic degradation), Face Similarity, and FID on VGGFace2 and CelebA-HQ. The samples cover four prompts and six comparison methods.

The anti-aesthetic score is negatively correlated with Face Similarity and positively correlated with FID, suggesting that lower preference-aligned quality often coincides with weaker identity preservation. This correlation motivates HAA but does not by itself rule out generic image degradation as a confounder. We therefore provide the controlled analyses in the following subsection to separate the proposed aesthetic mechanism from generic low-level distortion.

\subsection{Visualization Results} 
Fig.~\ref{fig:show} visualizes the perturbations generated by different protection methods on the CelebA-HQ dataset, complemented by quantitative imperceptibility metrics (LPIPS~\cite{zhang2018unreasonable} and SSIM~\cite{wang2004image}) to objectively evaluate the visual naturalness of protected images. Compared with Mist, CAAT, ASPL, and SimAC, the HAA method introduces fewer visually salient artifacts while still providing effective protection, indicating that it achieves a more favorable balance between preserving the natural appearance of shared images and reducing their utility for malicious customization.

Fig.~\ref{fig:visual} further compares the qualitative protection outcomes on VGGFace2 for two identities under four prompts. Across prompts, models fine-tuned on HAA-protected images tend to generate faces with substantially weaker identity-consistent details. In contrast, competing methods more often retain recognizable facial structure, which visually aligns with their higher identity leakage in our quantitative evaluation. Moreover, various figures in Appendix visually show the effectiveness of HAA in protecting individuals across different genders and ethnicities.

\subsection{Robustness Verification and Fault-Tolerant Mechanism}
We further evaluate HAA under challenging input conditions and discuss the fallback mechanism used when face detection fails.


\noindent\textbf{Robustness under extreme scenarios.}
We evaluate common input variations, including occlusion, rotation, and motion blur. As shown in Fig.~\ref{fig:show5}, HAA continues to reduce identity-consistent generation under these conditions, indicating that the perturbation remains effective beyond the standard clean-input setting.


\noindent \newtext{\textbf{Fallback for face-detection failure.}
LAA relies on valid face localization. Face detection succeeds in about 99\% of LAA updates on VGGFace2 and CelebA-HQ; when it fails, we use a central pseudo-box covering 64\% of the image area to keep the local branch differentiable, while GAA continues to provide full-image feedback. In the forced failure test in Fig.~\ref{fig:r2q3_failure}, adding GAA improves LAA at both 0\% and 60\% valid-local-detection availability. At 0\%, FDSR decreases from 0.719 to 0.562 and Face Similarity from 0.388 to 0.261; at 60\%, they decrease from 0.531 to 0.375 and from 0.299 to 0.152, respectively. Thus, GAA provides fallback anti-aesthetic feedback when local face supervision is absent or incomplete.}

\noindent \newtext{\textbf{Partial identity retention.}
HAA reduces identity evidence but does not theoretically guarantee complete identity removal. Residual identity can remain under weak perturbation budgets, after purification, under strong prompt mismatch or model mismatch, or in pipelines less dependent on pixel-level facial reconstruction. We therefore report both FDSR and ArcFace Face Similarity and include residual cases from Textual Inversion, DiffPure, SD-V3.0, and FLUX. HAA still lowers Face Similarity and preference-aligned quality in these settings, while GAA/LAA cooperation, pseudo-box fallback, and multi-image protection mitigate but do not eliminate such failures.}

\subsection{\newtext{Disentangling Aesthetic Degradation from Generic Image Degradation}}
\newtext{We further examine whether HAA's identity-removal effect is attributable to generic image degradation or to the proposed aesthetic mechanism. We use four complementary analyses: quantitative and qualitative LPIPS-controlled comparisons with generic degradations, a reward-control study, and ArcFace Grad-CAM visualization.}

\newtext{Tab.~\ref{tab:degradation_control} compares HAA with five generic degradations whose LPIPS values are all higher than HAA's (0.709--0.725 vs. 0.706). Despite larger perceptual distortion, these degradations retain substantially more identity evidence: Defocus Blur, Gaussian Blur, and Resolution Degradation still yield FDSR \(=1.000\), while Gaussian Noise and Salt-Pepper Noise retain much higher Face Similarity (0.482 and 0.476). By contrast, HAA reduces FDSR to 0.281 and Face Similarity to 0.117.}

\begin{table}[!t]
\centering
\caption{\newtext{Controlled comparison between generic image degradations and HAA under LPIPS-controlled perceptual distortion on CelebA-HQ with the prompt ``a photo of sks person''.}}
\label{tab:degradation_control}
\newtext{\resizebox{\linewidth}{!}{
\begin{tabular}{lccccc}
\toprule
Method & FDSR$\downarrow$ & Face Similarity$\downarrow$ & IR$\downarrow$ & FID$\uparrow$ & LPIPS \\
\midrule
Defocus Blur & 1.000 & 0.330 & -0.488 & 217.1 & 0.715 \\
Gaussian Blur & 1.000 & 0.383 & -0.265 & 215.8 & 0.722 \\
Resolution Degradation & 1.000 & 0.332 & 0.412 & 258.0 & 0.709 \\
Gaussian Noise & 0.719 & 0.482 & 0.449 & 153.8 & 0.725 \\
Salt-Pepper Noise & 0.469 & 0.476 & 0.587 & 135.8 & 0.709 \\
HAA & \textbf{0.281} & \textbf{0.117} & \textbf{-0.894} & \textbf{471.1} & 0.706 \\
\bottomrule
\end{tabular}
}}
\end{table}

\begin{figure}[!t]
\centering
\includegraphics[width=\linewidth]{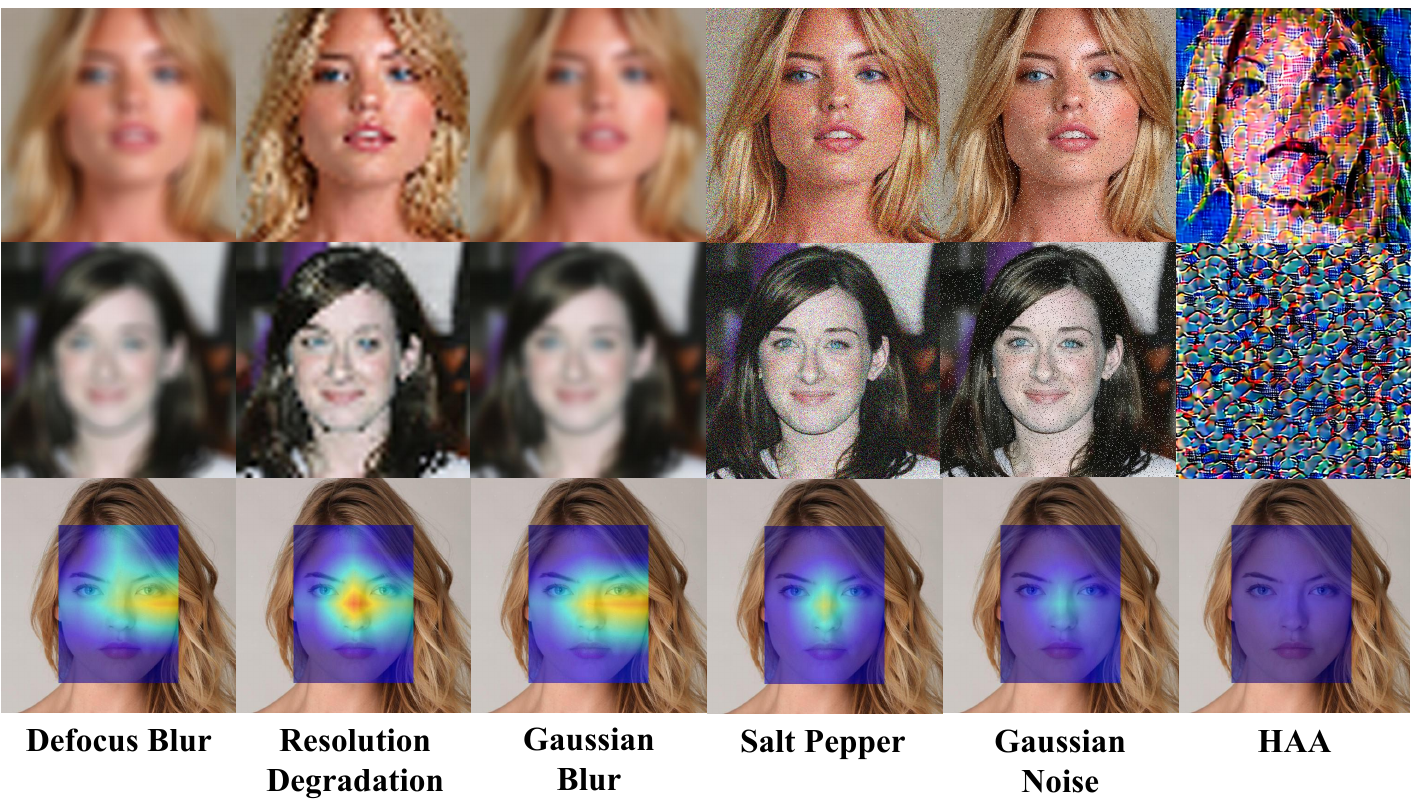}
\caption{\newtext{Qualitative LPIPS-controlled comparison and ArcFace Grad-CAM analysis. The first two rows compare generic degradations and HAA. The third row shows ArcFace embedding Grad-CAM responses with respect to the original identity; brighter regions indicate stronger identity evidence, while weaker responses indicate stronger de-identification.}}
\label{fig:degradation_gradcam}
\vspace{-10pt}
\end{figure}

\newtext{The first two rows of Fig.~\ref{fig:degradation_gradcam} provide a qualitative LPIPS-controlled comparison. Generic degradations introduce blur, resolution loss, or noise, but still preserve recognizable facial layouts and local identity details. HAA causes stronger identity-oriented disruption, consistent with Tab.~\ref{tab:degradation_control}.}

\newtext{Tab.~\ref{tab:reward_control} assesses aesthetic-mechanism-driven optimization. Replacing the aesthetic reward with a random reward does not improve over the baseline (FDSR 0.844 vs. 0.719; Face Similarity 0.394 vs. 0.388), whereas HAA achieves the best privacy metrics. Thus, meaningful aesthetic feedback, rather than a random reward signal, provides important evidence for disrupting identity-relevant cues.}

\begin{table}[!t]
\centering
\caption{\newtext{Reward-control study for aesthetic-mechanism-driven optimization. Clean denotes customization using unprotected original images; Baseline denotes protection without the proposed aesthetic reward terms;
Random replaces the aesthetic rewards with random signals under the same optimization protocol.}}
\label{tab:reward_control}
\newtext{\resizebox{0.8\linewidth}{!}{
\begin{tabular}{lcccc}
\toprule
Method & FDSR$\downarrow$ & Face Similarity$\downarrow$ & IR$\downarrow$ & FID$\uparrow$ \\
\midrule
Clean & 1.000 & 0.498 & 0.599 & 112.3 \\
Baseline & 0.719 & 0.388 & 0.183 & 321.6 \\
Random & 0.844 & 0.394 & 0.206 & 326.4 \\
HAA & \textbf{0.281} & \textbf{0.117} & \textbf{-0.894} & \textbf{471.1} \\
\bottomrule
\end{tabular}
}}
\end{table}

\newtext{The third row of Fig.~\ref{fig:degradation_gradcam} shows ArcFace Grad-CAM responses with respect to the original identity. Resolution Degradation remains bright around key facial regions, indicating residual original-identity evidence. HAA exhibits almost no high-response regions, indicating a larger distance from the original identity in ArcFace embedding space. Overall, generic degradation lacks an effective design for suppressing identity cues, whereas HAA achieves stronger de-identification through its aesthetics-driven mechanism.}

\begin{table}[!t]
\centering
\caption{Comparison of computational cost and VRAM requirements on CelebA-HQ.}
\label{tab:timecomparison}
\resizebox{0.95\linewidth}{!}{
\begin{tabular}{@{}c|cccc@{}}
\toprule
\textbf{Method} & \textbf{ VRAM (MiB)↓} & \textbf{Training Time (s)↓} & \textbf{FDSR↓} & \textbf{FID↑} \\ \hline
ASPL & 20417 & 152 & 0.750 & 359.5 \\
SimAC & 30327 & 327 & 0.438 & 388.9 \\
HAA & 30647 & 486 & 0.281 & 471.1 \\ \bottomrule
\end{tabular}}
\end{table}

\subsection{Comparison of computational cost and VRAM requirements on CelebA-HQ} 
Tab.~\ref{tab:timecomparison} compares computational cost and VRAM usage on CelebA-HQ. HAA consumes \(30647\) MiB VRAM, which is essentially on par with SimAC (\(30327\) MiB), suggesting that the hierarchical anti-aesthetic design does not materially increase the memory footprint. Although HAA increases training time from 327\,s to 486\,s, it delivers a noticeably larger gain in protection: FDSR decreases from \(0.438\) to \(0.281\) and FID increases from \(388.9\) to \(471.1\). Given our evaluation protocol, where lower FDSR and higher FID indicate stronger protection, these results suggest that HAA offers a favorable efficiency-effectiveness trade-off. It enhances privacy protection while incurring only modest additional VRAM usage and a relatively small increase in computational cost, which may be acceptable for resource-constrained deployments that prioritize protection strength.

\subsection{\newtext{Reward Model Validation}}
\newtext{Since \(RM_g\) and \(RM_l\) provide the optimization feedback for HAA, we validate them along three axes: effectiveness, identity specificity, and robustness/reliability.}

\noindent \newtext{\textbf{Aesthetic validity analysis.} We evaluate \(RM_g\) on ImageRewardDB for image-level aesthetic preference alignment and \(RM_l\) on SCUT-FBP5500 for face-local aesthetic preference alignment. As shown in Tables~\ref{tab:r2q2_rmg} and~\ref{tab:r2q2_rml}, \(RM_g\) provides meaningful global preference scores, and \(RM_l\) achieves the best preference accuracy among the compared facial scoring models.} \newtext{We further test downstream utility using the random-reward control in Tab.~\ref{tab:reward_control}. Replacing the aesthetic reward with a random signal does not improve protection over the baseline, whereas HAA substantially reduces FDSR and Face Similarity. Thus, HAA benefits from structured aesthetic feedback rather than arbitrary rewards.}

\begin{table}[!t]
\centering
\caption{\newtext{Validation of the global reward model \(RM_g\) on ImageRewardDB.}}
\label{tab:r2q2_rmg}
\newtext{\resizebox{\linewidth}{!}{
\begin{tabular}{lccccccc}
\toprule
\multirow{2}{*}{Model} & \multirow{2}{*}{Pref. Acc.} & \multicolumn{3}{c}{Recall} & \multicolumn{3}{c}{Filter} \\
\cmidrule(lr){3-5} \cmidrule(lr){6-8}
 &  & @1 & @2 & @4 & @1 & @2 & @4 \\
\midrule
CLIP & 54.82 & 27.22 & 48.52 & 78.17 & 29.65 & 51.75 & 76.82 \\
Aesthetic & 57.35 & 30.73 & 53.91 & 75.74 & 32.08 & 54.45 & 76.55 \\
BLIP & 57.76 & 30.73 & 50.67 & 77.63 & 33.42 & 56.33 & 80.59 \\
\(RM_g\) & 64.43 & 43.60 & 66.80 & 94.40 & 48.80 & 72.80 & 95.20 \\
\bottomrule
\end{tabular}}}
\vspace{-10pt}
\end{table}

\begin{table}[!t]
\centering
\caption{\newtext{Validation of the local facial reward model \(RM_l\) on SCUT-FBP5500.}}
\label{tab:r2q2_rml}
\newtext{\resizebox{\linewidth}{!}{
\begin{tabular}{lccccccc}
\toprule
\multirow{2}{*}{Model} & \multirow{2}{*}{Pref. Acc.} & \multicolumn{3}{c}{Recall} & \multicolumn{3}{c}{Filter} \\
\cmidrule(lr){3-5} \cmidrule(lr){6-8}
 &  & @1 & @2 & @4 & @1 & @2 & @4 \\
\midrule
CLIP & 53.63 & 15.31 & 30.69 & 59.64 & 16.62 & 30.69 & 59.42 \\
Aesthetic & 55.78 & 17.35 & 29.20 & 54.55 & 26.69 & 43.82 & 67.71 \\
BLIP & 59.33 & 20.25 & 40.69 & 71.31 & 18.11 & 33.71 & 59.85 \\
\(RM_l\) & 67.20 & 35.56 & 59.64 & 87.67 & 22.69 & 41.09 & 68.65 \\
\bottomrule
\end{tabular}}}
\vspace{-10pt}
\end{table}

\begin{figure}[!t]
\centering
\newtext{
\includegraphics[width=\linewidth]{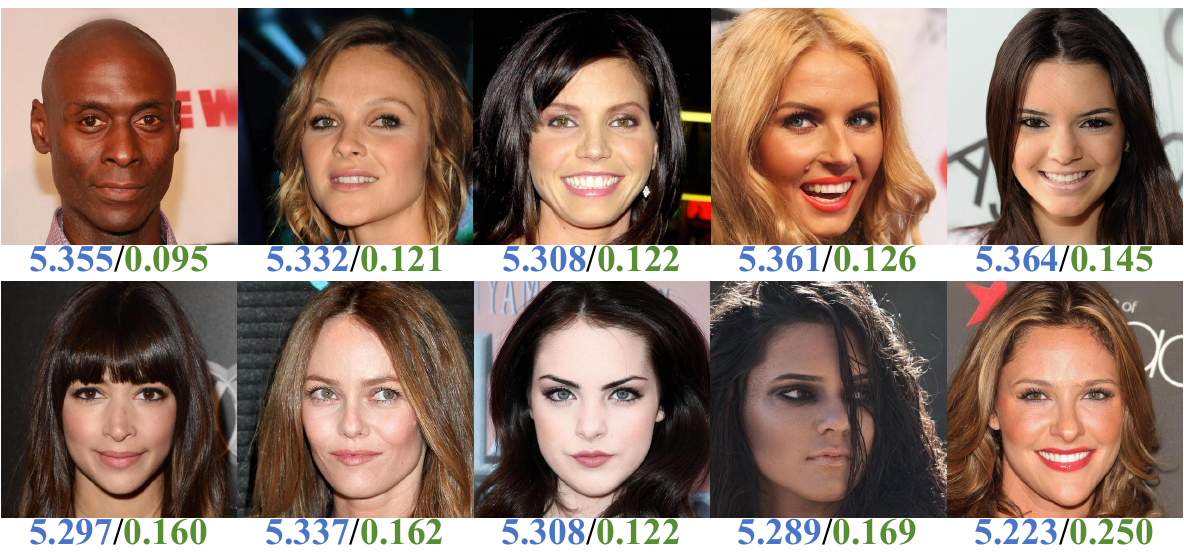}
\vspace{2pt}
\includegraphics[width=\linewidth]{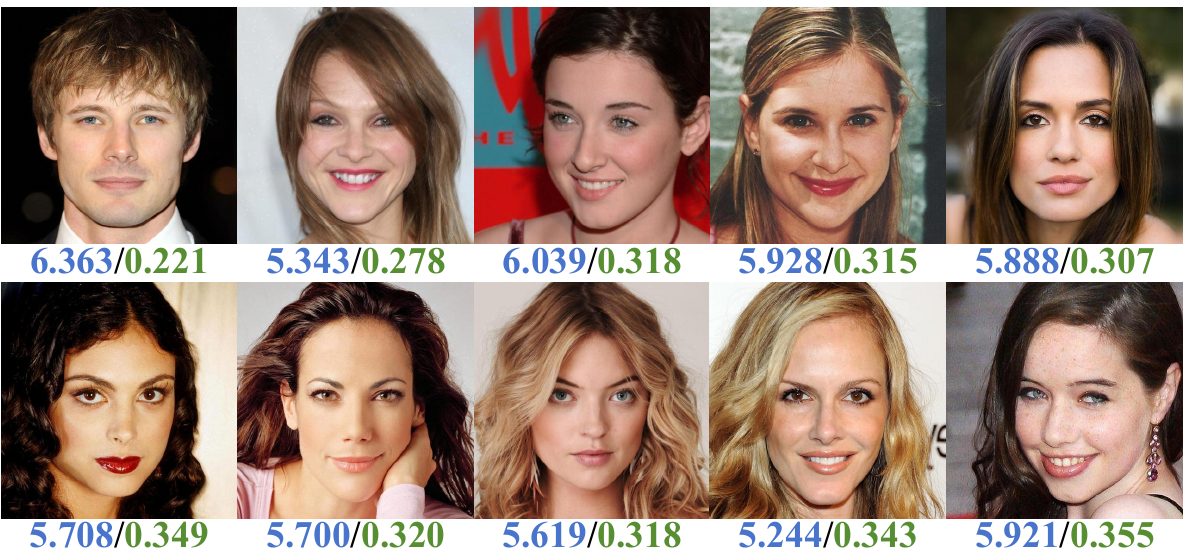}
\caption{Identity-specificity diagnostics of \(RM_l\). Top: different identities with similar aesthetic quality, where each tile reports the \(RM_l\) score / absolute difference from the Aesthetic Scorer. Bottom: same-identity groups under background and expression changes, where each tile reports the mean / standard deviation.}
\label{fig:r2q2_identity_cases}
}
\vspace{-10pt}
\end{figure}

\noindent \newtext{\textbf{Identity-specificity analysis.} \(RM_l\) is not trained or used as an identity classifier; it only scores detected facial regions for local aesthetics. As shown in Fig.~\ref{fig:r2q2_identity_cases}, different identities with similar aesthetic quality receive close scores, with a small average absolute score difference of about \(0.157\) compared to the Aesthetic Scorer~\cite{schuhmann2022laion}. Images of the same identity under different backgrounds and expressions also yield relatively small score variation (standard deviation \(0.221\)--\(0.355\)), indicating stable scoring across non-identity variations. These diagnostics indicate that \(RM_l\)'s scores do not rely on specific identity labels, but are mainly driven by local facial aesthetic quality.}

\begin{figure}[!t]
\centering
\includegraphics[width=0.9\linewidth]{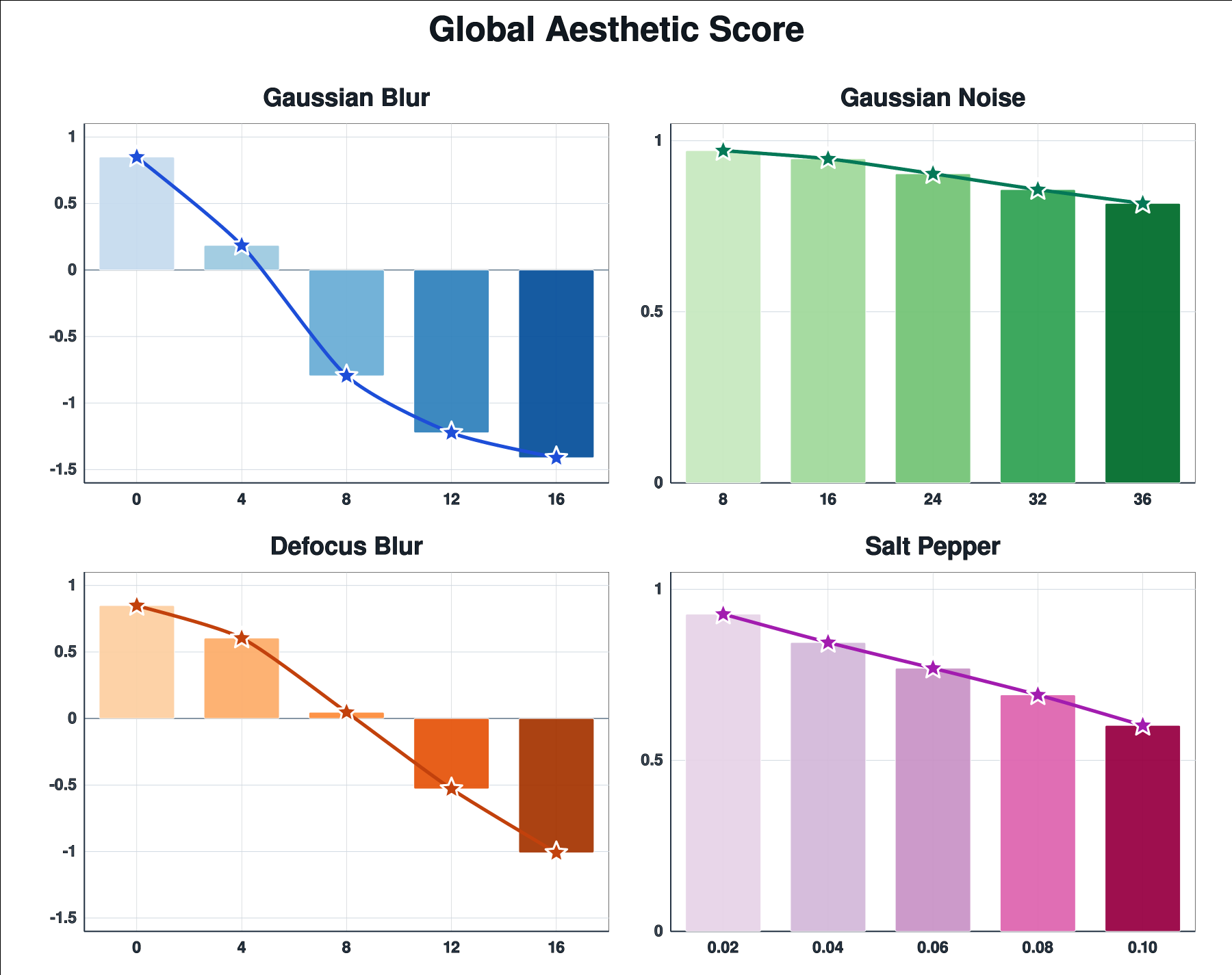}
\caption{\newtext{Robustness analysis of the global reward model \(RM_g\). Under Gaussian Blur, Gaussian Noise, Defocus Blur, and Salt-Pepper Noise, the global aesthetic score consistently decreases as degradation intensity increases.}}
\label{fig:r2q2_aes_global}
\end{figure}

\begin{figure}[!t]
\centering
\includegraphics[width=0.9\linewidth]{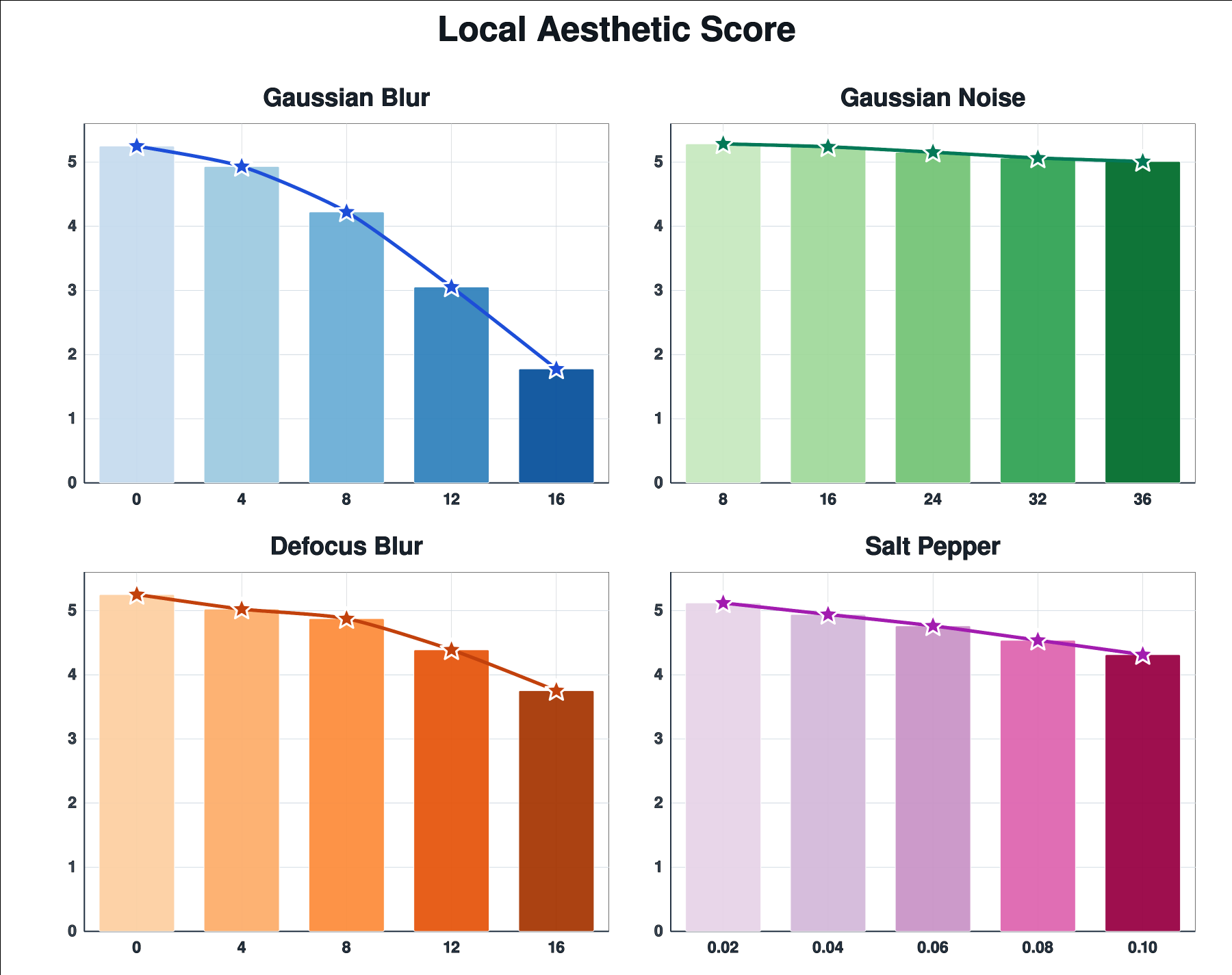}
\caption{\newtext{Robustness analysis of the local reward model \(RM_l\). Under the same four degradation types, the face-local aesthetic score consistently decreases with stronger degradation.}}
\label{fig:r2q2_aes_local}
\vspace{-10pt}
\end{figure}

\noindent \newtext{\textbf{Robustness and reliability analysis.} We apply Gaussian Blur, Gaussian Noise, Defocus Blur, and Salt-Pepper Noise with increasing intensities. As shown in Figs.~\ref{fig:r2q2_aes_global} and~\ref{fig:r2q2_aes_local}, both \(RM_g\) and \(RM_l\) consistently assign lower scores as degradation intensity increases, indicating stable and quality-sensitive feedback.} \newtext{Overall, these results evaluate the reward models at the scorer, downstream-optimization, and diagnostic-behavior levels, supporting their use as frozen preference scorers in HAA.}

\subsection{More details}
\newtext{Both reward models use a BLIP-style preference-scorer architecture with a ViT-L image encoder, a 12-layer transformer text encoder, cross-modal fusion, and an MLP scalar head. For \(RM_g\), we use the public ImageRewardDB dataset for prompt-conditioned image-level rankings. The preference data contain 8,878 prompts and 136,892 pairwise comparisons, constructed from ranked groups of 4--9 generated images for each prompt. The reported preference-accuracy test split contains 466 prompts and 6,399 comparison pairs, while Recall/Filter are evaluated on another 371 prompts with 8 images per prompt; the remaining annotated prompts are used for training. Given a prompt \(C_g\), a preferred image \(x^w\), and a less preferred image \(x^l\), the global scorer is optimized with a pairwise logistic ranking loss, i.e., it is encouraged to satisfy \(RM_g(x^w,C_g)>RM_g(x^l,C_g)\). This makes \(RM_g\) sensitive to global human-preference factors such as prompt alignment, fidelity, and overall aesthetics.}

\newtext{For \(RM_l\), we construct face-level preference pairs and apply the same preference-scorer principle at the face level. Facial regions are detected from LAION natural images, masked, and inpainted with a diffusion inpainting pipeline to create degraded face regions. The original face crop is treated as \(x^w_l\), while the inpainted/degraded face is treated as \(x^l_l\), yielding about 46k face-quality pairs from 23k natural images. The local prompt is fixed to the face-level prompt \(C_l\), so the scorer focuses on facial plausibility and aesthetic detail rather than full-image context. \(RM_l\) is optimized with the same pairwise logistic ranking objective, encouraging \(RM_l(x^w_l,C_l)>RM_l(x^l_l,C_l)\). No identity labels are used in this training; the supervision is quality/preference ordering. After training, both \(RM_g\) and \(RM_l\) are frozen and only provide differentiable feedback for HAA perturbation optimization.} \newtext{We also report the training protocol: 70\% of BLIP Transformer layers are frozen, the learning rate is set to \(1\times 10^{-5}\), the batch size is 64, a cosine learning-rate schedule is used, and training is conducted on four RTX 5880 Ada GPUs. Training can also be conducted on a single RTX 5880 Ada GPU, although this increases the training time.}

\section{Conclusion}
Existing anti-customization methods often overlook key aesthetic cues, limiting their ability to remove identities. To address this issue, we propose a novel Hierarchical Anti-Aesthetics (HAA) framework, which includes both Global Anti-Aesthetics and Local Anti-Aesthetics branches. These branches work together to leverage aesthetic cues from global to local and encourage anti-alignment with human aesthetic preferences, thereby reducing customized DMs' ability to reconstruct facial details. Extensive experiments show that HAA outperforms existing methods and provides a useful tool for protecting facial privacy and copyright.

\bibliographystyle{IEEEtran}
\bibliography{ref}

@String(CVPR= {IEEE Conf. Comput. Vis. Pattern Recog.})

@String(ICME = {Int. Conf. Multimedia and Expo})

@String(ICASSP=	{ICASSP})

@String(AAAI = {AAAI})

@String(CVPR  = {CVPR})

@String(ICME  =	{ICME})

@misc{labs2025flux1kontextflowmatching,
      title={FLUX.1 Kontext: Flow Matching for In-Context Image Generation and Editing in Latent Space},
      author={Black Forest Labs and Stephen Batifol and Andreas Blattmann and Frederic Boesel and Saksham Consul and Cyril Diagne and Tim Dockhorn and Jack English and Zion English and Patrick Esser and Sumith Kulal and Kyle Lacey and Yam Levi and Cheng Li and Dominik Lorenz and Jonas Müller and Dustin Podell and Robin Rombach and Harry Saini and Axel Sauer and Luke Smith},
      year={2025},
      eprint={2506.15742},
      archivePrefix={arXiv},
      primaryClass={cs.GR},
      url={https://arxiv.org/abs/2506.15742},
}

@inproceedings{esser2024scaling,
  title={Scaling rectified flow transformers for high-resolution image synthesis},
  author={Esser, Patrick and Kulal, Sumith and Blattmann, Andreas and Entezari, Rahim and M{\"u}ller, Jonas and Saini, Harry and Levi, Yam and Lorenz, Dominik and Sauer, Axel and Boesel, Frederic and others},
  booktitle={Forty-first international conference on machine learning},
  year={2024}
}

@inproceedings{deng2019arcface,
  title={Arcface: Additive angular margin loss for deep face recognition},
  author={Deng, Jiankang and Guo, Jia and Xue, Niannan and Zafeiriou, Stefanos},
  booktitle={Proceedings of the IEEE/CVF conference on computer vision and pattern recognition},
  pages={4690--4699},
  year={2019}
}

@inproceedings{deng2020retinaface,
  title={Retinaface: Single-shot multi-level face localisation in the wild},
  author={Deng, Jiankang and Guo, Jia and Ververas, Evangelos and Kotsia, Irene and Zafeiriou, Stefanos},
  booktitle={Proceedings of the IEEE/CVF conference on computer vision and pattern recognition},
  pages={5203--5212},
  year={2020}
}

@article{yu2022surprising,
  title={The surprising effectiveness of ppo in cooperative multi-agent games},
  author={Yu, Chao and Velu, Akash and Vinitsky, Eugene and Gao, Jiaxuan and Wang, Yu and Bayen, Alexandre and Wu, Yi},
  journal={Advances in neural information processing systems},
  volume={35},
  pages={24611--24624},
  year={2022}
}

@inproceedings{frank2020leveraging,
  title={Leveraging frequency analysis for deep fake image recognition},
  author={Frank, Joel and Eisenhofer, Thorsten and Sch{\"o}nherr, Lea and Fischer, Asja and Kolossa, Dorothea and Holz, Thorsten},
  booktitle={International conference on machine learning},
  pages={3247--3258},
  year={2020},
  organization={PMLR}
}

@inproceedings{yang2015facial,
  title={From facial parts responses to face detection: A deep learning approach},
  author={Yang, Shuo and Luo, Ping and Loy, Chen-Change and Tang, Xiaoou},
  booktitle={Proceedings of the IEEE international conference on computer vision},
  pages={3676--3684},
  year={2015}
}

@misc{rombach2021highresolution,
      title={High-Resolution Image Synthesis with Latent Diffusion Models}, 
      author={Robin Rombach and Andreas Blattmann and Dominik Lorenz and Patrick Esser and Björn Ommer},
      year={2021},
      eprint={2112.10752},
      archivePrefix={arXiv},
      primaryClass={cs.CV}
}

@article{tinio2011image,
  title={Image quality and the aesthetic judgment of photographs: Contrast, sharpness, and grain teased apart and put together.},
  author={Tinio, Pablo PL and Leder, Helmut and Strasser, Marlies},
  journal={Psychology of Aesthetics, Creativity, and the Arts},
  volume={5},
  number={2},
  pages={165},
  year={2011},
  publisher={Educational Publishing Foundation}
}

@article{palmer2013visual,
  title={Visual aesthetics and human preference},
  author={Palmer, Stephen E and Schloss, Karen B and Sammartino, Jonathan},
  journal={Annual review of psychology},
  volume={64},
  number={1},
  pages={77--107},
  year={2013},
  publisher={Annual Reviews}
}

@article{xu2024imagereward,
  title={Imagereward: Learning and evaluating human preferences for text-to-image generation},
  author={Xu, Jiazheng and Liu, Xiao and Wu, Yuchen and Tong, Yuxuan and Li, Qinkai and Ding, Ming and Tang, Jie and Dong, Yuxiao},
  journal={Advances in Neural Information Processing Systems},
  volume={36},
  year={2024}
}

@article{gallego2022personalizing,
  title={Personalizing text-to-image generation via aesthetic gradients},
  author={Gallego, Victor},
  journal={arXiv preprint arXiv:2209.12330},
  year={2022}
}

@article{wu2024vmix,
  title={VMix: Improving Text-to-Image Diffusion Model with Cross-Attention Mixing Control},
  author={Wu, Shaojin and Ding, Fei and Huang, Mengqi and Liu, Wei and He, Qian},
  journal={arXiv preprint arXiv:2412.20800},
  year={2024}
}

@inproceedings{zhu2022celebv,
  title={CelebV-HQ: A large-scale video facial attributes dataset},
  author={Zhu, Hao and Wu, Wayne and Zhu, Wentao and Jiang, Liming and Tang, Siwei and Zhang, Li and Liu, Ziwei and Loy, Chen Change},
  booktitle={European conference on computer vision},
  pages={650--667},
  year={2022},
  organization={Springer}
}

@inproceedings{cao2018vggface2,
  title={Vggface2: A dataset for recognising faces across pose and age},
  author={Cao, Qiong and Shen, Li and Xie, Weidi and Parkhi, Omkar M and Zisserman, Andrew},
  booktitle={2018 13th IEEE international conference on automatic face \& gesture recognition (FG 2018)},
  pages={67--74},
  year={2018},
  organization={IEEE}
}

@article{heusel2017gans,
  title={Gans trained by a two time-scale update rule converge to a local nash equilibrium},
  author={Heusel, Martin and Ramsauer, Hubert and Unterthiner, Thomas and Nessler, Bernhard and Hochreiter, Sepp},
  journal={Advances in neural information processing systems},
  volume={30},
  year={2017}
}

@inproceedings{ruiz2023DreamBooth,
  title={Dreambooth: Fine tuning text-to-image diffusion models for subject-driven generation},
  author={Ruiz, Nataniel and Li, Yuanzhen and Jampani, Varun and Pritch, Yael and Rubinstein, Michael and Aberman, Kfir},
  booktitle={Proceedings of the IEEE/CVF conference on computer vision and pattern recognition},
  pages={22500--22510},
  year={2023}
}

@inproceedings{kumari2023multi,
  title={Multi-concept customization of text-to-image diffusion},
  author={Kumari, Nupur and Zhang, Bingliang and Zhang, Richard and Shechtman, Eli and Zhu, Jun-Yan},
  booktitle={Proceedings of the IEEE/CVF Conference on Computer Vision and Pattern Recognition},
  pages={1931--1941},
  year={2023}
}

@article{ramesh2022hierarchical,
  title={Hierarchical text-conditional image generation with clip latents},
  author={Ramesh, Aditya and Dhariwal, Prafulla and Nichol, Alex and Chu, Casey and Chen, Mark},
  journal={arXiv preprint arXiv:2204.06125},
  volume={1},
  number={2},
  pages={3},
  year={2022}
}

@article{ho2020denoising,
  title={Denoising diffusion probabilistic models},
  author={Ho, Jonathan and Jain, Ajay and Abbeel, Pieter},
  journal={Advances in neural information processing systems},
  volume={33},
  pages={6840--6851},
  year={2020}
}

@article{song2020denoising,
  title={Denoising diffusion implicit models},
  author={Song, Jiaming and Meng, Chenlin and Ermon, Stefano},
  journal={arXiv preprint arXiv:2010.02502},
  year={2020}
}

@inproceedings{rombach2022high,
  title={High-resolution image synthesis with latent diffusion models},
  author={Rombach, Robin and Blattmann, Andreas and Lorenz, Dominik and Esser, Patrick and Ommer, Bj{\"o}rn},
  booktitle={Proceedings of the IEEE/CVF conference on computer vision and pattern recognition},
  pages={10684--10695},
  year={2022}
}

@article{podell2023sdxl,
  title={Sdxl: Improving latent diffusion models for high-resolution image synthesis},
  author={Podell, Dustin and English, Zion and Lacey, Kyle and Blattmann, Andreas and Dockhorn, Tim and M{\"u}ller, Jonas and Penna, Joe and Rombach, Robin},
  journal={arXiv preprint arXiv:2307.01952},
  year={2023}
}

@inproceedings{radford2021learning,
  title={Learning transferable visual models from natural language supervision},
  author={Radford, Alec and Kim, Jong Wook and Hallacy, Chris and Ramesh, Aditya and Goh, Gabriel and Agarwal, Sandhini and Sastry, Girish and Askell, Amanda and Mishkin, Pamela and Clark, Jack and others},
  booktitle={International conference on machine learning},
  pages={8748--8763},
  year={2021},
  organization={PMLR}
}

@inproceedings{xu2024perturbing,
  title={Perturbing Attention Gives You More Bang for the Buck: Subtle Imaging Perturbations That Efficiently Fool Customized Diffusion Models},
  author={Xu, Jingyao and Lu, Yuetong and Li, Yandong and Lu, Siyang and Wang, Dongdong and Wei, Xiang},
  booktitle={Proceedings of the IEEE/CVF Conference on Computer Vision and Pattern Recognition},
  pages={24534--24543},
  year={2024}
}

@inproceedings{wang2024simac,
  title={SimAC: a simple anti-customization method for protecting face privacy against text-to-image synthesis of diffusion models},
  author={Wang, Feifei and Tan, Zhentao and Wei, Tianyi and Wu, Yue and Huang, Qidong},
  booktitle={Proceedings of the IEEE/CVF Conference on Computer Vision and Pattern Recognition},
  pages={12047--12056},
  year={2024}
}

@inproceedings{van2023anti,
  title={Anti-dreambooth: Protecting users from personalized text-to-image synthesis},
  author={Van Le, Thanh and Phung, Hao and Nguyen, Thuan Hoang and Dao, Quan and Tran, Ngoc N and Tran, Anh},
  booktitle={Proceedings of the IEEE/CVF International Conference on Computer Vision},
  pages={2116--2127},
  year={2023}
}

@article{liang2023adversarial,
  title={Adversarial example does good: Preventing painting imitation from diffusion models via adversarial examples},
  author={Liang, Chumeng and Wu, Xiaoyu and Hua, Yang and Zhang, Jiaru and Xue, Yiming and Song, Tao and Xue, Zhengui and Ma, Ruhui and Guan, Haibing},
  journal={arXiv preprint arXiv:2302.04578},
  year={2023}
}

@inproceedings{bar2022text2live,
  title={Text2live: Text-driven layered image and video editing},
  author={Bar-Tal, Omer and Ofri-Amar, Dolev and Fridman, Rafail and Kasten, Yoni and Dekel, Tali},
  booktitle={European conference on computer vision},
  pages={707--723},
  year={2022},
  organization={Springer}
}

@inproceedings{kim2022diffusionclip,
  title={Diffusionclip: Text-guided diffusion models for robust image manipulation},
  author={Kim, Gwanghyun and Kwon, Taesung and Ye, Jong Chul},
  booktitle={Proceedings of the IEEE/CVF Conference on Computer Vision and Pattern Recognition},
  pages={2426--2435},
  year={2022}
}

@inproceedings{lugmayr2022repaint,
  title={Repaint: Inpainting using denoising diffusion probabilistic models},
  author={Lugmayr, Andreas and Danelljan, Martin and Romero, Andres and Yu, Fisher and Timofte, Radu and Van Gool, Luc},
  booktitle={Proceedings of the IEEE/CVF Conference on Computer Vision and Pattern Recognition},
  pages={11461--11471},
  year={2022}
}

@article{ho2022cascaded,
  title={Cascaded diffusion models for high fidelity image generation},
  author={Ho, Jonathan and Saharia, Chitwan and Chan, William and Fleet, David J and Norouzi, Mohammad and Salimans, Tim},
  journal={Journal of Machine Learning Research},
  volume={23},
  number={47},
  pages={1--33},
  year={2022}
}

@inproceedings{han2023svdiff,
  title={Svdiff: Compact parameter space for diffusion fine-tuning},
  author={Han, Ligong and Li, Yinxiao and Zhang, Han and Milanfar, Peyman and Metaxas, Dimitris and Yang, Feng},
  booktitle={Proceedings of the IEEE/CVF International Conference on Computer Vision},
  pages={7323--7334},
  year={2023}
}

@article{gal2022image,
  title={An image is worth one word: Personalizing text-to-image generation using textual inversion},
  author={Gal, Rinon and Alaluf, Yuval and Atzmon, Yuval and Patashnik, Or and Bermano, Amit H and Chechik, Gal and Cohen-Or, Daniel},
  journal={arXiv preprint arXiv:2208.01618},
  year={2022}
}

@article{zheng2023understanding,
  title={Understanding and Improving Adversarial Attacks on Latent Diffusion Model},
  author={Zheng, Boyang and Liang, Chumeng and Wu, Xiaoyu and Liu, Yan},
  journal={arXiv preprint arXiv:2310.04687},
  year={2023}
}

@inproceedings{gu2022vector,
  title={Vector quantized diffusion model for text-to-image synthesis},
  author={Gu, Shuyang and Chen, Dong and Bao, Jianmin and Wen, Fang and Zhang, Bo and Chen, Dongdong and Yuan, Lu and Guo, Baining},
  booktitle={Proceedings of the IEEE/CVF conference on computer vision and pattern recognition},
  pages={10696--10706},
  year={2022}
}

@article{vice2024bagm,
  title={Bagm: A backdoor attack for manipulating text-to-image generative models},
  author={Vice, Jordan and Akhtar, Naveed and Hartley, Richard and Mian, Ajmal},
  journal={IEEE Transactions on Information Forensics and Security},
  year={2024},
  publisher={IEEE}
}

@inproceedings{huang2024personalization,
  title={Personalization as a shortcut for few-shot backdoor attack against text-to-image diffusion models},
  author={Huang, Yihao and Juefei-Xu, Felix and Guo, Qing and Zhang, Jie and Wu, Yutong and Hu, Ming and Li, Tianlin and Pu, Geguang and Liu, Yang},
  booktitle={Proceedings of the AAAI Conference on Artificial Intelligence},
  volume={38},
  number={19},
  pages={21169--21178},
  year={2024}
}

@article{seow2022comprehensive,
  title={A comprehensive overview of Deepfake: Generation, detection, datasets, and opportunities},
  author={Seow, Jia Wen and Lim, Mei Kuan and Phan, Rapha{\"e}l CW and Liu, Joseph K},
  journal={Neurocomputing},
  volume={513},
  pages={351--371},
  year={2022},
  publisher={Elsevier}
}

@article{lu2015rating,
  title={Rating image aesthetics using deep learning},
  author={Lu, Xin and Lin, Zhe and Jin, Hailin and Yang, Jianchao and Wang, James Z},
  journal={IEEE Transactions on Multimedia},
  volume={17},
  number={11},
  pages={2021--2034},
  year={2015},
  publisher={IEEE}
}

@inproceedings{zhang2018unreasonable,
  title={The unreasonable effectiveness of deep features as a perceptual metric},
  author={Zhang, Richard and Isola, Phillip and Efros, Alexei A and Shechtman, Eli and Wang, Oliver},
  booktitle={Proceedings of the IEEE conference on computer vision and pattern recognition},
  pages={586--595},
  year={2018}
}

@article{wang2004image,
  title={Image quality assessment: from error visibility to structural similarity},
  author={Wang, Zhou and Bovik, Alan C and Sheikh, Hamid R and Simoncelli, Eero P},
  journal={IEEE transactions on image processing},
  volume={13},
  number={4},
  pages={600--612},
  year={2004},
  publisher={IEEE}
}

@inproceedings{xu2025harnessing,
  title     = {Harnessing Global-Local Collaborative Adversarial Perturbation for Anti-Customization},
  author    = {Xu, Long and Wang, Jiakai and Hao, Haojie and Qin, Haotong and Zhao, Jiejie and Liu, Xianglong},
  booktitle = {Proceedings of the IEEE/CVF Conference on Computer Vision and Pattern Recognition (CVPR)},
  pages     = {13414--13423},
  year      = {2025},
  doi       = {10.1109/CVPR52734.2025.01252}
}

@article{schuhmann2022laion,
  title={Laion-5b: An open large-scale dataset for training next generation image-text models},
  author={Schuhmann, Christoph and Beaumont, Romain and Vencu, Richard and Gordon, Cade and Wightman, Ross and Cherti, Mehdi and Coombes, Theo and Katta, Aarush and Mullis, Clayton and Wortsman, Mitchell and others},
  journal={Advances in neural information processing systems},
  volume={35},
  pages={25278--25294},
  year={2022}
}

@article{wang2025fast,
  title={Fast adversarial training with weak-to-strong spatial-temporal consistency in the frequency domain on videos},
  author={Wang, Songping and Liu, Hanqing and Lyu, Yueming and Hu, Xiantao and He, Ziwen and Wang, Wei and Shan, Caifeng and Wang, Liang},
  journal={IEEE Transactions on Information Forensics and Security},
  volume={21},
  pages={681--696},
  year={2025},
  publisher={IEEE}
}

@article{wang2026exposing,
  title={Exposing and Defending the Achilles' Heel of Video Mixture-of-Experts},
  author={Wang, Songping and Liu, Qinglong and Lyu, Yueming and Li, Ning and He, Ziwen and Shan, Caifeng},
  journal={arXiv preprint arXiv:2602.01369},
  year={2026}
}

@inproceedings{wang2026runawayevil,
  title={Runawayevil: Jailbreaking the image-to-video generative models},
  author={Wang, Songping and Qian, Rufan and Lyu, Yueming and Liu, Qinglong and Zou, Linzhuang and Qin, Jie and Liu, Songhua and Shan, Caifeng},
  booktitle={Proceedings of the IEEE/CVF Conference on Computer Vision and Pattern Recognition},
  pages={9296--9305},
  year={2026}
}

@inproceedings{wang2024public,
  title={Public-domain locator for boosting attack transferability on videos},
  author={Wang, Songping and Liu, Hanqing and Zhao, Haochen},
  booktitle={2024 IEEE International Conference on Multimedia and Expo (ICME)},
  pages={1--6},
  year={2024},
  organization={IEEE}
}

@article{wang2025exploring,
  title={Exploring adversarial transferability between kolmogorov-arnold networks},
  author={Wang, Songping and Yue, Xinquan and Lyu, Yueming and Shan, Caifeng},
  journal={arXiv preprint arXiv:2503.06276},
  year={2025}
}

@article{wang2025anti,
  title={Anti-aesthetics: Protecting facial privacy against customized text-to-image synthesis},
  author={Wang, Songping and Lyu, Yueming and Liu, Shiqi and Li, Ning and Tong, Tong and Sun, Hao and Shan, Caifeng},
  journal={arXiv preprint arXiv:2504.12129},
  year={2025}
}

@inproceedings{wang2024effective,
  title={An effective end-to-end solution for multimodal action recognition},
  author={Wang, Songping and Rao, Haoxiang and Hu, Xiantao and Lyu, Yueming and Shan, Caifeng},
  booktitle={International Conference on Pattern Recognition},
  pages={324--338},
  year={2024},
  organization={Springer}
}

@article{wei2023efficient,
  title={Efficient robustness assessment via adversarial spatial-temporal focus on videos},
  author={Wei, Xingxing and Wang, Songping and Yan, Huanqian},
  journal={IEEE Transactions on Pattern Analysis and Machine Intelligence},
  volume={45},
  number={9},
  pages={10898--10912},
  year={2023},
  publisher={IEEE}
}

@incollection{meng2023coarse,
  title={Coarse to fine segmentation method enables accurate and efficient segmentation of organs and tumor in abdominal ct},
  author={Meng, Hui and Zhao, Haochen and Yang, Deqian and Wang, Songping and Li, Zhenpeng},
  booktitle={MICCAI Challenge on Fast and Low-Resource Semi-supervised Abdominal Organ Segmentation},
  pages={115--129},
  year={2023},
  publisher={Springer}
}

@article{li2026agentcanary,
  title={AgentCanary: A Security Evaluation Framework for Autonomous AI Agents in Real Executable Environments},
  author={Li, Peiyang and Wang, Songping and Huang, Yi and Shi, Yanhua and Zhang, Chenhao and Li, Qi and Lyu, Yueming and Shan, Caifeng and Li, Fengting and Feng, Chao and others},
  journal={arXiv preprint arXiv:2606.10484},
  year={2026}
}

@inproceedings{chen2026neurerase,
  title={NeurErase: Selective Deactivation of Neurons for Erasing Concepts in Diffusion Models},
  author={Chen, Ziyuan and Lyu, Yueming and Meng, Zheling and Rao, Haoxiang and Li, Ning and Wang, Songping and Shan, Caifeng},
  booktitle={ICASSP 2026-2026 IEEE International Conference on Acoustics, Speech and Signal Processing (ICASSP)},
  pages={14077--14081},
  year={2026},
  organization={IEEE}
}

@inproceedings{jin2025frequency,
  title={Frequency Domain Distributed Perturbations: Towards Query-Efficient Black-Box Adversarial Video Attack},
  author={Jin, Teng and He, Ziwen and Fu, Zhangjie and Wang, Songping and Lyu, Yueming and Shi, Yufei},
  booktitle={Proceedings of the 33rd ACM International Conference on Multimedia},
  pages={8360--8368},
  year={2025}
}

@article{zhou2025multimodal,
  title={Multimodal Video-Based Heart Rate Estimation with Temporal Difference Transformer},
  author={Zhou, Zhiqin and Wang, Songping and Shan, Caifeng},
  year={2025}
}
\end{document}